\pdfoutput=1

\documentclass[11pt]{article}

\usepackage[final]{acl}

\usepackage{times}
\usepackage{latexsym}

\usepackage[T1]{fontenc}

\usepackage[utf8]{inputenc}

\usepackage{microtype}

\usepackage{inconsolata}

\usepackage{graphicx}

\usepackage{booktabs}
\usepackage{amsmath}
\usepackage{multirow}
\usepackage{makecell}
\usepackage{amsfonts}
\usepackage{bbm}
\usepackage{colortbl}

\usepackage{xcolor}
\usepackage{subcaption}

\definecolor{OliveGreen}{HTML}{3C8031}
\definecolor{PineGreen}{HTML}{008B72}

\DeclareMathOperator*{\argmax}{argmax}

\title{A Systematic Analysis of Base Model Choice for Reward Modeling}

\author{
 \textbf{Kian Ahrabian\textsuperscript{1,2}}\and
 \textbf{Pegah Jandaghi\textsuperscript{1,2}}\and
 \textbf{Negar Mokhberian\textsuperscript{1,2}}\AND
 \textbf{Sai Praneeth Karimireddy\textsuperscript{1}}\and
 \textbf{Jay Pujara\textsuperscript{1,2}}\AND
 {\normalfont \textsuperscript{1}University of Southern California, Los Angeles, USA}\\
 \textsuperscript{2}Information Sciences Institute, Marina del Rey, USA\\
 \texttt{\{ahrabian,jandaghi,nmokhber,karimire\}@usc.edu}, \texttt{jpujara@isi.edu} \\
}

\begin{document}
\maketitle

\begin{abstract}
Reinforcement learning from human feedback (RLHF) and, at its core, reward modeling have become a crucial part of training powerful large language models (LLMs).
One commonly overlooked factor in training high-quality reward models (RMs) is the effect of the base model, which is becoming more challenging to choose given the rapidly growing pool of LLMs.
In this work, we present a systematic analysis of the effect of base model selection on reward modeling performance.
Our results show that the performance can be improved by up to 14\% compared to the most common (\textit{i.e.,} default) choice.
Moreover, we showcase the strong statistical relation between some existing benchmarks and downstream performances.
We also demonstrate that the results from a small set of benchmarks could be combined to boost the model selection ($+$18\% on average in the top 5-10).
Lastly, we illustrate the impact of different post-training steps on the final performance and explore using estimated data distributions to reduce performance prediction error.
\end{abstract}

\section{Introduction}
\label{sec:intro}

\begin{figure}[t]
  \includegraphics[width=\columnwidth]{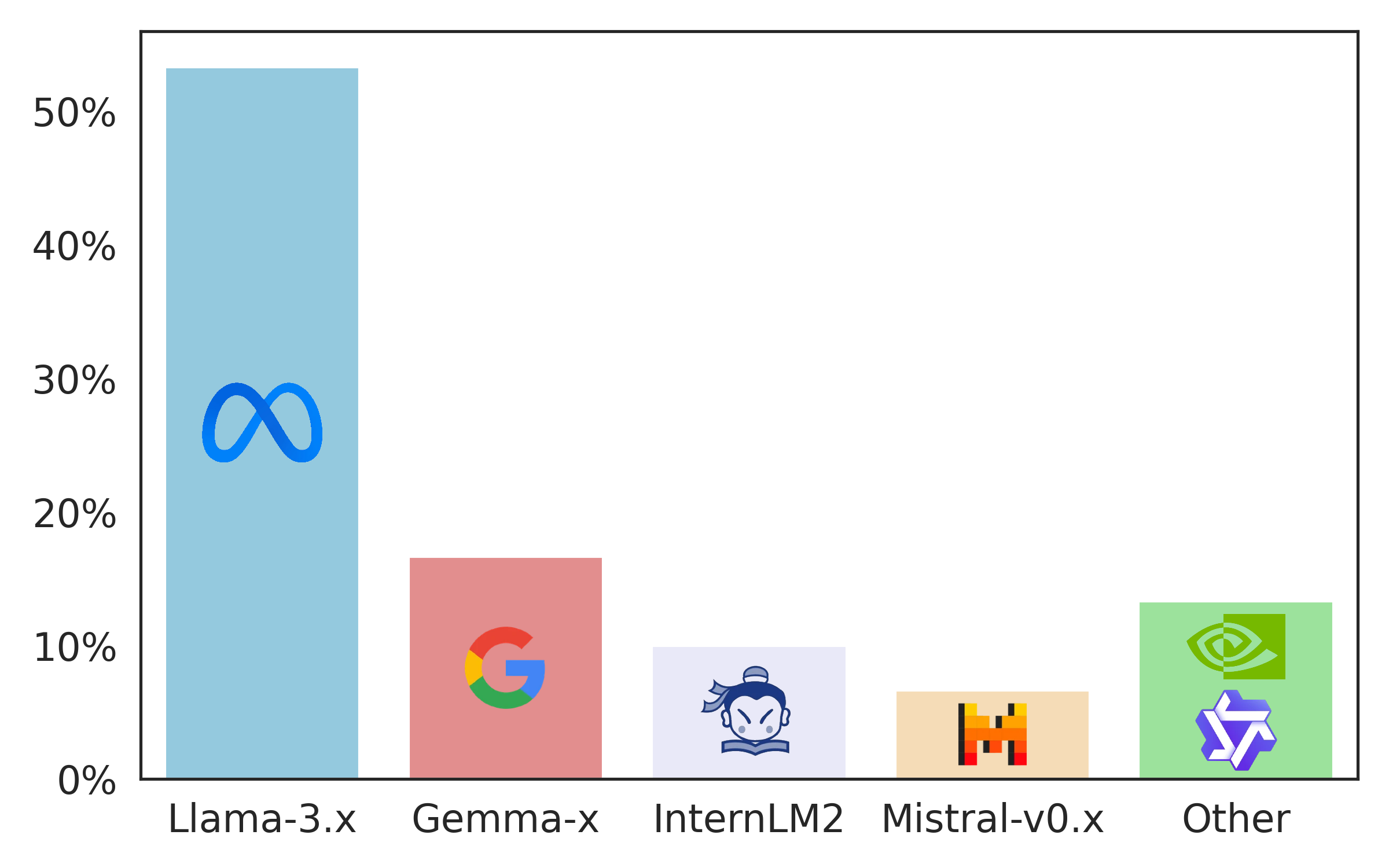}
  \caption{\textbf{Ratio of the base models used in the top 30 entries of RewardBench (Dec 2024).} Almost all the entries are trained on top of a small set of base models (\textit{e.g.,} Llama-3.x models comprise 50\% of the entries).}
  \label{fig:ratio}
\end{figure}

Reinforcement learning from human feedback (RLHF)~\cite{stiennon2020learning,ouyang2022training,bai2022constitutional} has been a critical part of recent advancements in large language models (LLMs) such as OpenAI's O1~\cite{openai_o1}, Anthropic's Claude~\cite{anthropic_claude}, and Google's Gemini~\cite{team2023gemini}.
At the core of RLHF methods, Reward Models (RMs) are used to guide the LLM (\textit{i.e.,} policy) training by scoring generated responses~\cite{schulman2017proximal,ahmadian2024back}.
Most commonly, RMs are evaluated on RewardBench\footnote{\href{https://huggingface.co/spaces/allenai/reward-bench}{allenai/reward-bench}}~\cite{lambert2024rewardbench}, consisting of 2985 binary preference tasks, 23 subtasks, and four subcategories.
The RewardBench leaderboard reflects a bias toward a single model family, with more than 50\% of the top 30 entries (see \autoref{fig:ratio}) built on top of a Llama-3.x model~\cite{dubey2024llama}
However, relying on a single model family without exploration is inherently suboptimal, regardless of Llama-3.x models' quality.

Considering this suboptimality, we hypothesize that the base model is a critical hyperparameter that substantially impacts the downstream performance.
To test this hypothesis, we compare 40 popular models across various sizes and families (see \autoref{app:model} for more details).
Our experiments show that replacing the popular base model (\textit{i.e.,} LLama-3.x) with the best model of similar size leads to gains ranging from 3\% to 14\%.
While these results prove our hypothesis, running such a search over the plethora of available models is extremely expensive.
This obstacle inspires the need for robust approaches that could either limit the search perimeter or help us make a selection apriori.
However, the criteria for selecting a model apriori are often unclear and multifaceted.

Prior works in RLHF~\cite{stiennon2020learning,gao2023scaling} have examined the relation between the model size and performance.
Moreover, recent works~\cite{ruan2024observational,polo2024sloth} have used compute metrics (\textit{e.g.,} training tokens) and simple capabilities measured by standard benchmarks (\textit{e.g.,} MMLU~\cite{hendrycks2021measuring}) to predict emergent capabilities of LLMs.
Inspired by these works, we use these features to systematically analyze the base models to identify core capabilities and attributes that yield high-quality RMs.
Our experiments show that while performances on many benchmarks and reward modeling have strong statistical correlations, they are insufficient for the broader model selection problem.
Moreover, we show significant improvements ($+$18\% on average in the top 5-10) can be gained over any single benchmark-based selection, only using a small subset of benchmarks.

While our analysis covers various elements, it does not investigate the effect of different training stages of a model, which have grown in numbers with recent advancements.
Hence, we separately investigate the pre-training and post-training stages, relying on publicly available intermediate checkpoints~\cite{lambert2024tulu3}.
For the post-training stage, we demonstrate the positive impact of the supervised fine-tuning (SFT) stage ($+$15.5\%) while showcasing the negative effect of the following alignment steps (3-5\% drop).
For the pre-training stage, we focus on estimating~\cite{bakman2024mars} and analyzing the data composition, which has emerged as a key driving factor in recent developments~\cite{abdin2024phi,abdin2024phi4,yang2024qwen2}.
Our experiments show estimated distributions' variability across model families, which we use to reduce our regression model's error ($+$1.5\%).

To summarize, our contributions are as follows:
\begin{itemize}
    \item We showcase the significance of the base model choice, which could improve upon the most common (\textit{i.e.,} default) choice up to 14\% in a size-controlled setting.
    \item We analyze the statistical relation between performances on standard benchmarks and reward modeling, showcasing strong correlations (Pearson $\ge$ 0.8) on many while illustrating their shortcoming in model selection (\textit{i.e.,} small overlap on top models)
    \item We show a simple performance prediction regression model based on benchmarks' results, when employed for model selection, can achieve $+$18\% overlap on average over the top 5-10, compared to the benchmark with the highest correlation.
    \item We showcase the positive impact of the post-training stages, especially SFT, achieving up to $+$15.5\% gains on publicly available models.
    Moreover, we expose the negative impact of the standard post-SFT alignment steps, leading to a 3-5\% performance drop.
    \item We exhibit the potential of using estimated data distributions, which improves our regression model's performance by $+$1.5\%.
\end{itemize}

\section{Related Work}
\label{sec:related}
\paragraph{Reward Modeling}

Recently, there has been a lot of effort in crafting better training datasets~\cite{liu2024skywork,wang2024helpsteer2-p} and improving training architectures~\cite{dorka2024quantile,lou2024uncertainty,zhang2024general,wang2024interpretable}.
However, the core objective for reward modeling still revolves around two main approaches: Bradley-Terry w/ Binary Preferences~\cite{ziegler2019fine,bradley1952rank} and Regression w/ Multi-Attribute Scores~\cite{wang2023helpsteer} (see~\autoref{sec:rm} for more details).
For datasets, RMs are commonly trained on labeled preference datasets such as UltraFeedback~\cite{cui2023ultrafeedback}, HelpSteer2~\cite{wang2024helpsteer2}, and Magpie~\cite{xu2024magpie}.

\paragraph{Reward Model Evaluation}

Until recently, one of the biggest challenges of training RMs has been evaluating the trained models in isolation.
The lack of test sets in the released datasets made evaluation difficult without going through the highly costly policy training step.
To overcome this issue, recent works~\cite{lambert2024rewardbench,liu2024rm,gureja2024m} have introduced standardized benchmarks for evaluating these models.
Among these benchmarks, RewardBench~\cite{lambert2024rewardbench} is the most popular, with more than 150 entries at the time of writing this article.



\section{Reward Modeling}
\label{sec:rm}

\begin{figure*}[t]
  \begin{subfigure}[t]{\textwidth}
    \includegraphics[width=\textwidth]{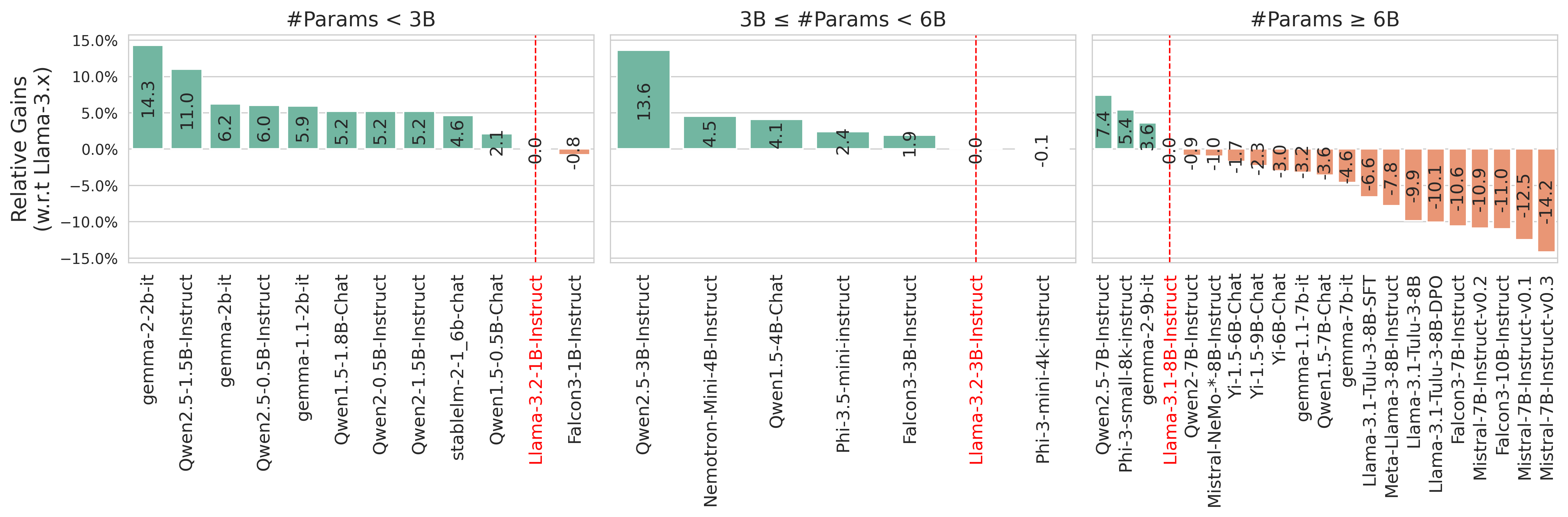}
    \caption{\textbf{Bradley-Terry w/ Binary Preferences}}
    \vspace{0.3cm}
  \end{subfigure}
  \begin{subfigure}[t]{\textwidth}
    \includegraphics[width=\textwidth]{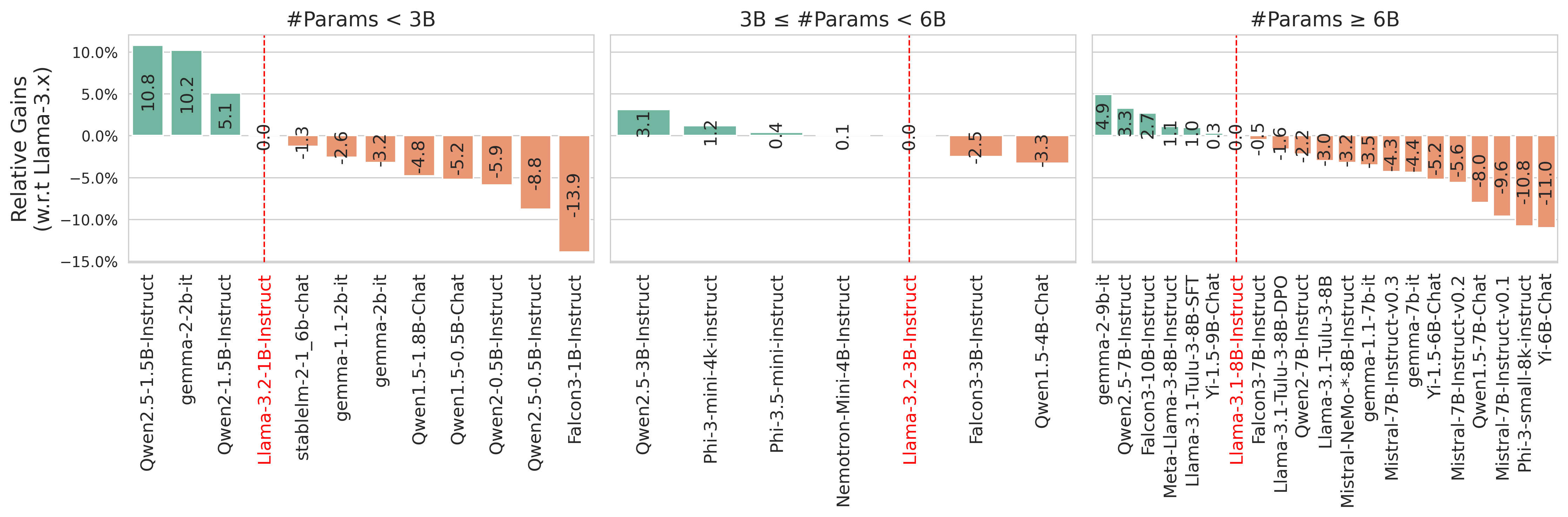}
    \caption{\textbf{Regression w/ Multi-Attribute Scores}}
  \end{subfigure}
  \caption{\textbf{Reward Modeling Performance Gains.} Relative gains are illustrated concerning the Llama-3.x model (marked as red) within the same group.}
  \label{fig:diff}
\end{figure*}



\subsection{Training}

\paragraph{Models.}
For our experiments, we use 40 different chat models from prominent publishers such as Microsoft, Google, and Meta, with sizes ranging from 494M to 10.30B (\textit{i.e.,} the largest model we could train on our cluster).
\autoref{app:model} provides more details on these models.
 
\paragraph{Bradley-Terry w/ Binary Preferences.}
The most popular choice for reward modeling is the Bradley-Terry (BT)~\cite{bradley1952rank,ziegler2019fine} model.
The underlying assumption of BT is that for a pair of responses $\mathcal{Y} = (y_1, y_2)$, the human preference distribution $\rho^*$ is generated from a latent reward function $r^*(x, y)$, which we only have indirect access to.
This assumption can be formalized as
\begin{equation}
    \rho^*(y_1 \succ y_2 | x) = \frac{\exp(r^*(x, y_1))}{\sum_y^{\mathcal{Y}}\exp(r^*(x, y))} \  .
\end{equation}
Then, framing BT as a binary classification task, we can parameterize the reward function and optimize a negative log-likelihood loss as
\begin{gather}
    \resizebox{0.896\columnwidth}{!}{$\mathcal{L}_{BT} = -\mathbb{E}_{(x,y_w,y_l)\sim\mathcal{D}}\left[ \log \sigma(\zeta(x, y_w) - \zeta(x, y_l)) \right]$}
\end{gather}
where $\mathcal{D} = \{(x^i,y_w^i,y_k^i)\}_{i=1}^N \sim \rho^*$ is a binary preferences dataset and $\zeta$ is an LLM with a linear head that outputs a single scalar as the reward value.

To create a compatible dataset, first, an LLM $\xi$ generates pairs of responses for samples from a given prompt dataset $\mathcal{D}_x$:
\begin{equation}
    \mathcal{D}_\xi = \{(x,y_1,y_2) | \{y_1,y_2\}\sim\xi(x)\}_{x\sim\mathcal{D}_x} \  .
\end{equation}
Then, the pairs are labeled by humans (or synthetically) to obtain the binary preferences:
\begin{equation}
   \mathcal{D} = \{ (x, y_w, y_l) | (y_w \succ y_l; x) \}_{(x, y_1, y_2)\sim\mathcal{D}_\xi} \  .
\end{equation}

We follow a similar setup for training the reward models as~\citet{wang2024helpsteer2-p}.
Specifically, each model is trained for one epoch on the HelpSteer2-Preference dataset, using a global batch size of 64, a constant learning rate, searched over $\{5,6,7,8,9\}e-7 \cup \{1,2,3,4,5\}e-6$ for each model separately, and an AdamW optimizer~\cite{loshchilov2017decoupled} with 20 warm-up steps.
Each model is saved every 20 steps, and the final model is chosen based on the accuracy of the saved models on the validation set.

\paragraph{Regression w/ Multi-Attribute Scores.}
While less explored compared to BT, Regression reward models~\cite{wang2023helpsteer,wang2024interpretable,wang2024helpsteer2} have been posting impressive performance recently, topping the RewardBench at multiple points (\textit{e.g.,} \texttt{ArmoRM-Llama3-8B-v0.1}\footnote{\href{https://huggingface.co/RLHFlow/ArmoRM-Llama3-8B-v0.1}{RLHFlow/ArmoRM-Llama3-8B-v0.1}} and \texttt{Nemotron-4-340B-Reward}\footnote{\href{https://huggingface.co/nvidia/Nemotron-4-340B-Reward}{nvidia/Nemotron-4-340B-Reward}}).
In contrast to the binary preferences, each sample is annotated with multiple values along different attributes (\textit{e.g.,} Coherence, Correctness, Verbosity, etc.).
Then, given an input $x$, an output score vector $y \in \mathbb{R}^n$, and an LLM $\phi$, we optimize
\begin{align}
    \mathcal{L}_{R} &= \text{MSE}(\phi(x)^{(-1)}W_\phi, y)
\end{align}
where $\phi(x)^{(-1)} \in \mathbb{R}^{\text{dim}(\phi)}$ is the last hidden state and $W_\phi \in \mathbb{R}^{\text{dim}(\phi) \times n}$ is a trainable linear projection (\textit{i.e.,} a linear layer).
This formulation leads to more flexible and interpretable reward models.
To train the models, we follow a similar setup as~\citet{wang2024helpsteer2}.
Specifically, each model is trained for two epochs on the HelpSteer2 dataset, using a global batch size of 64, a constant learning rate, searched over $\{1,3,5,7,9\}e-\{6,7\}$ for each model separately, and an AdamW optimizer with 20 warm-up steps.
Since RewardBench only supports BT models, for each model, we search for an optimal merge vector, $w_m$, as
\begin{gather}
    \psi(x) = (\phi(x)^{(-1)}W_\phi))^Tw \\
    w_m = \argmax_{w \in S} \sum_{x_c,x_r}^D \mathbbm{1}\left(\psi(x_c) > \psi(x_r)\right)
\end{gather}
where $D$ is the validation set of HelpSteer2-Preference~\cite{wang2024helpsteer2-p}, $x_c$ and $x_r$ are chosen and rejected responses, respectively, and $S = \{0.05k\}_{k=0,\dots,20}^4 \times \{-0.05k\}_{k=0,\dots,20}$ ($\sim$4M combinations).
We follow the approach in \texttt{Nemotron-4-340B-Reward} to assign positive weights for \textit{Helpfulness}, \textit{Correctness}, \textit{Coherence}, \textit{Complexity}, and a negative weight for \textit{Verbosity}.
Finally, we pick the model with the highest validation performance.

\subsection{Evaluation}
Following prior work~\cite{wang2024helpsteer2,wang2024helpsteer2-p,dorka2024quantile,lou2024uncertainty,zhang2024general,wang2024interpretable} and due to its popularity (\textit{e.g.,} more than 150 entries), we evaluate our trained models using RewardBench~\cite{lambert2024rewardbench}, which contains $\sim$3k assorted tasks from 23 different datasets.
Each task consists of a binary preference sample and is categorized into one of the following four categories: \textit{Chat}, \textit{Chat-Hard}, \textit{Safety}, and \textit{Reasoning}.
We report the accuracy for each category and an overall score by averaging the accuracies.

\subsection{Experimental Results}
To make a fairer comparison, we partition the models into three groups, each representing a range of roughly 3B parameters: $\{< 3\text{B}, (\ge3\text{B},<6\text{B}), \ge 6\text{B}\}$.
Then, we calculate the relative gains concerning the Llama-3.x model for each group (\textit{i.e.,} the default choice) within the same group.
\autoref{fig:diff} present our results models trained using Bradley-Terry (w/ binary preferences) and Regression (w/ multi-attribute scores).
While Llama-3.x models perform exceptionally well across our experiments, within each group, a few models post superior performances, with margins up to $\sim$14\%.
Specifically, looking at these top performances, models from the Qwen2.5 and Gemma-2 families consistently improve upon the results of their Llama-3.x counterpart, presenting reliable alternatives.
Moreover, these experiments showcase the potentially high variances in performance within groups of models with similar sizes, which, in many cases, is the main limiting factor for model selection.

\begin{figure*}[t]
  \begin{subfigure}[t]{\textwidth}
    \includegraphics[width=\textwidth]{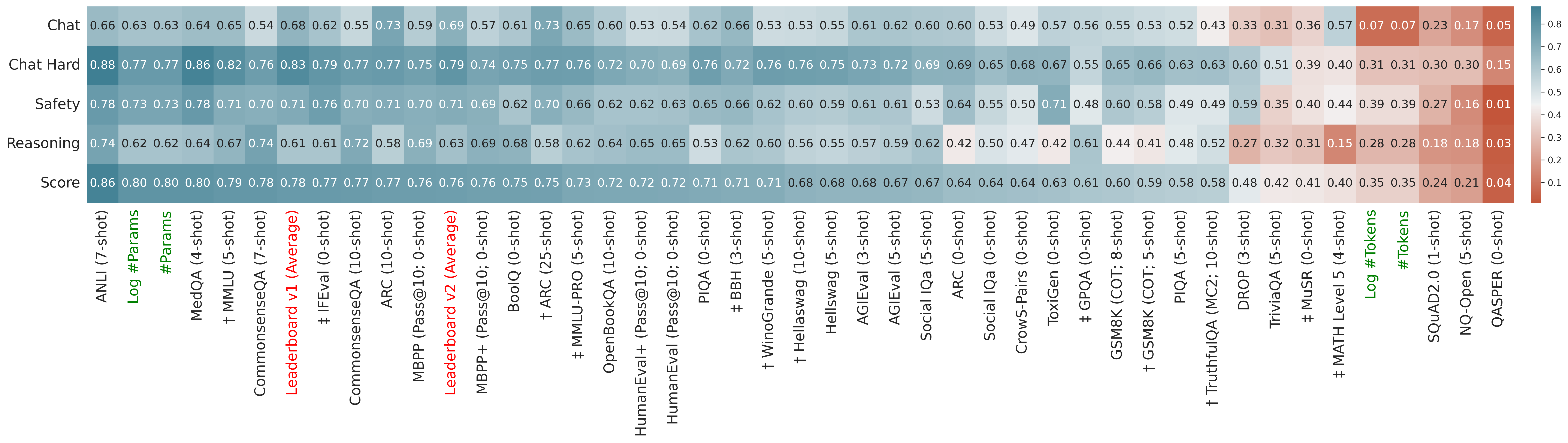}
    \caption{\textbf{Spearman Correlation}}
    \vspace{0.3cm}
  \end{subfigure}
  \begin{subfigure}[t]{\textwidth}
    \includegraphics[width=\textwidth]{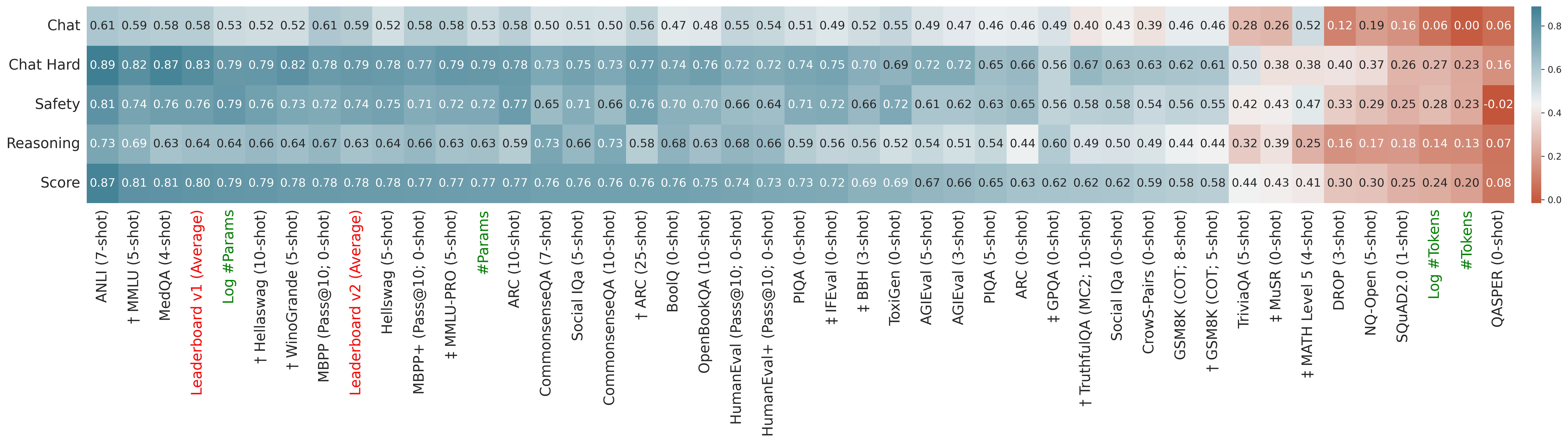}
    \caption{\textbf{Pearson Correlation}}
  \end{subfigure}
  \caption{\textbf{Statistical Correlation w.r.t. Reward Modeling Performance.} The subset benchmarks of Open LLM Leaderboard v2 (v1) are denoted with an $\ddagger$ ($\dagger$). \textit{Text Colors:} \textcolor{red}{Red} $\rightarrow$ Aggregate benchmark, \textcolor{OliveGreen}{Green} $\rightarrow$ Training metric.}
  \label{fig:reg}
\end{figure*}

\section{Benchmarks as Latent Skills Proxies}
\label{sec:bench}

\subsection{Statistical Correlation}
\paragraph{Setup.}
Practitioners often test their models on various benchmarks, covering many topics such as reasoning, coding, etc.
These benchmarks, along with aggregate benchmarks such as Open LLM Leaderboard~\cite{open-llm-leaderboard-v1,open-llm-leaderboard-v2} and HELM~\cite{liang2022holistic}, act as a proxy measurement of the true capabilities of LLMs.
Consequently, many of them are often used for model selection.
For our analysis, we curate a list of 33 common benchmarks as reported in Llama-3.x~\cite{dubey2024llama}, Gemma-2~\cite{team2024gemma}, Phi-3.x~\cite{abdin2024phi}, and Qwen2.5~\cite{yang2024qwen2} families (see \autoref{app:data} for more details).
Besides these benchmarks, we also include training metrics such as the number of parameters and the number of training tokens, as they are commonly used in formulating scaling laws~\cite{ruan2024observational,polo2024sloth}.

\paragraph{Results.}
\autoref{fig:reg} presents our correlation analysis between these benchmarks/metrics and the final reward modeling performances\footnote{On the \textit{Chat} subcategory, all the models achieve s $90\text{-}95\%$ performance, which makes them challenging to distinguish considering minor performance variances; hence, we observe relatively low correlations across benchmarks.}.
As evident, some benchmarks showcase a very high ($\ge 0.8$) correlation, both on Pearson and Spearman, with ANLI~\cite{williams-etal-2022-anlizing} consistently beating other benchmarks across different subcategories.

\paragraph{Significance Test.}
We test the significance of the correlation coefficient with the following statistic:
\begin{equation}
    t_c = \frac{r\sqrt{n - 2}}{\sqrt{1 - r^2}}
\end{equation}
where $r$ is the sample correlation coefficient, and $n$ is the sample size, which leads to a threshold $t_c$ of 0.316 ($n=40$) for $p$-value < 0.05.
Using this threshold, we observe that most of the benchmarks' correlations have statistical significance.

\begin{figure}[t]
  \includegraphics[width=\columnwidth]{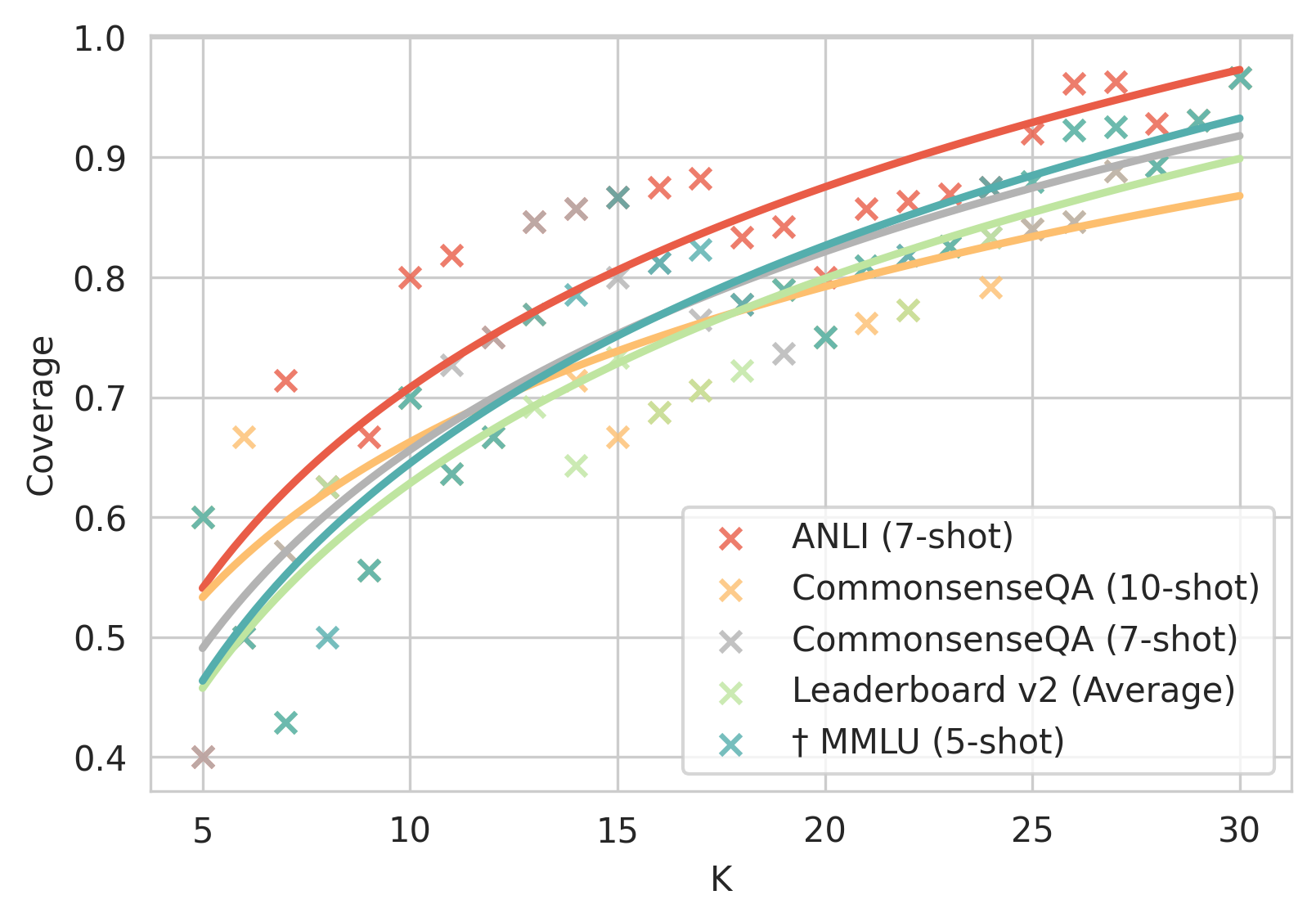}
  \caption{\textbf{Benchmark's Coverage.} We only retain benchmarks with at least 0.4 and 0.7 coverage at $k=5$ and $k=10$, respectively.}
  \label{fig:top_cov}
\end{figure}

\paragraph{Coverage Test.}
While a high correlation shows a strong statistical relationship between the two variables, we also care about the coverage at different points in their rankings.
Given a benchmark $\beta$ and reward bench $\rho$, we formally define the coverage at top-$k$ as
\begin{equation}
   \mathcal{C}(\beta, \rho, \mathcal{L}, k) = \frac{| \mathcal{T}_\beta(\mathcal{L}, k) \cap \mathcal{T}_\rho(\mathcal{L}, k) |}{k}
\end{equation}
where $\mathcal{L}$ is a set of LLMs and $\mathcal{T}_x(y, z)$ is the top $z$ LLMs in $y$ on benchmark $x$.
To simulate a real-world search where we need high coverage at higher ranks, we filter out any benchmark with less than 0.4 and 0.7 coverage at $k=5$ and $k=10$, respectively.
\autoref{fig:top_cov} illustrates the coverage values from $k=5$ to $k=30$ on the remaining benchmarks (see \autoref{app:data} for more details).
Notably, all the benchmarks mostly follow a log-linear coverage pattern concerning $k$, with ANLI outperforming the other benchmarks.
However, we also observe a relatively low coverage at higher ranks, which mitigates the effectiveness of using these benchmarks for model selection.

\begin{figure}[t]
  \includegraphics[width=\columnwidth]{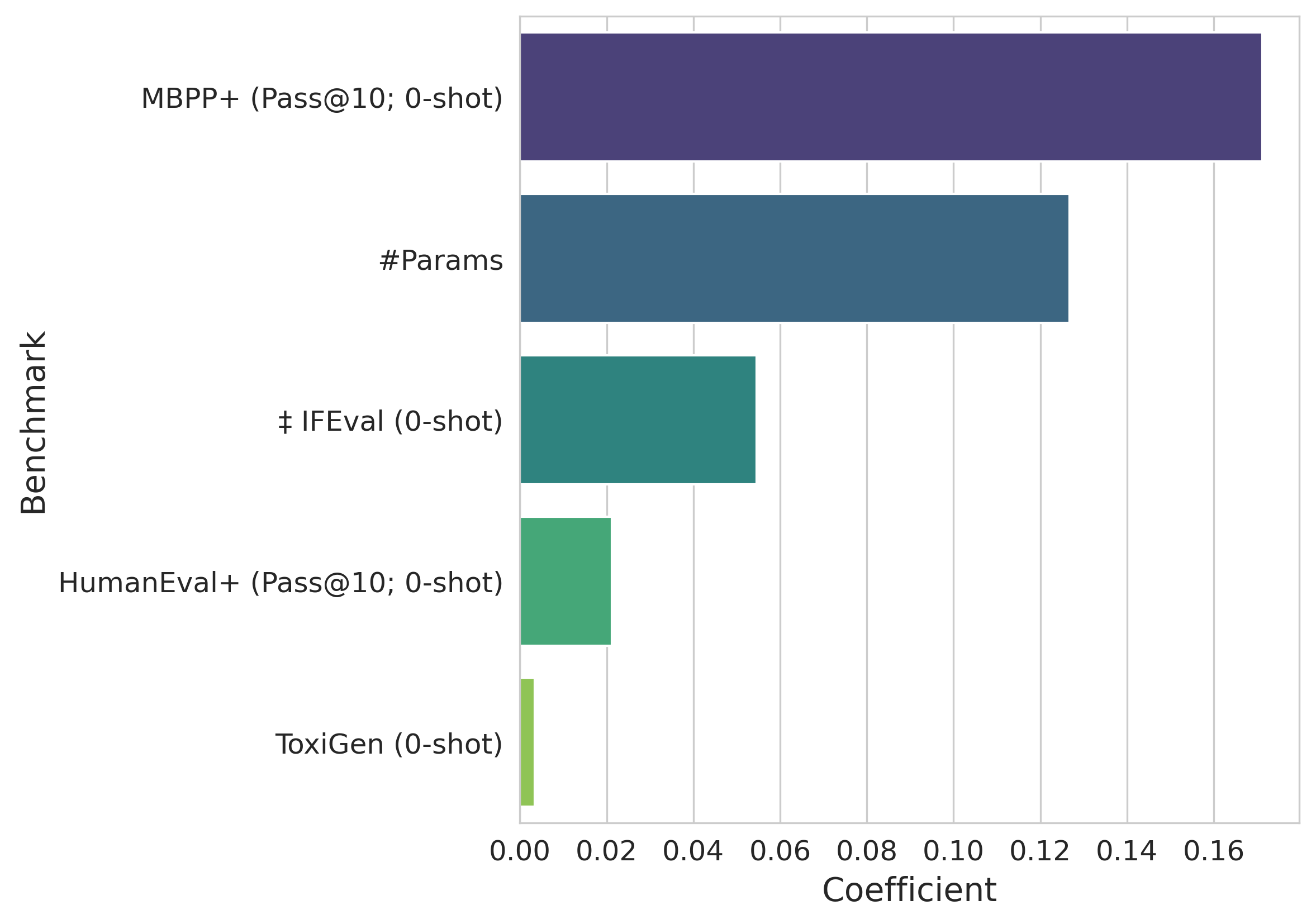}
  \caption{\textbf{Coefficients.} Only five benchmarks are assigned a non-zero weight by the trained model. The topics of these benchmarks are as follows: \textit{Coding} $\rightarrow$ MBPP+ and HumanEval+, \textit{Safety} $\rightarrow$ ToxiGen, \textit{General} $\rightarrow$ IFEval, and \textit{Training Metrics} $\rightarrow$ \#Params (see \autoref{app:data} for more details).}
  \label{fig:elast_coeff}
\end{figure}

\subsection{Regression Analysis}
\label{sec:bench:reg}
\paragraph{Setup.}
Considering the aforementioned low coverage in single-benchmark model selection, we hypothesize that combining the performances from a small set of benchmarks will yield much better predictive performance.
To test this hypothesis, we run a 10-fold cross-validation experiment on an Elastic Net model, searching over the following hyperparameters: degree $\in \{1, 2, 3\}$, $\alpha \in \{0.1, 0.01, 0.001, 0.0001\}$, and l1\_ratio $\in \{0.0, 0.25, 0.5, 0.75, 1.0\}$.
Then, we fit a model over all samples using the best hyperparameters.

\begin{figure}[t]
  \includegraphics[width=\columnwidth]{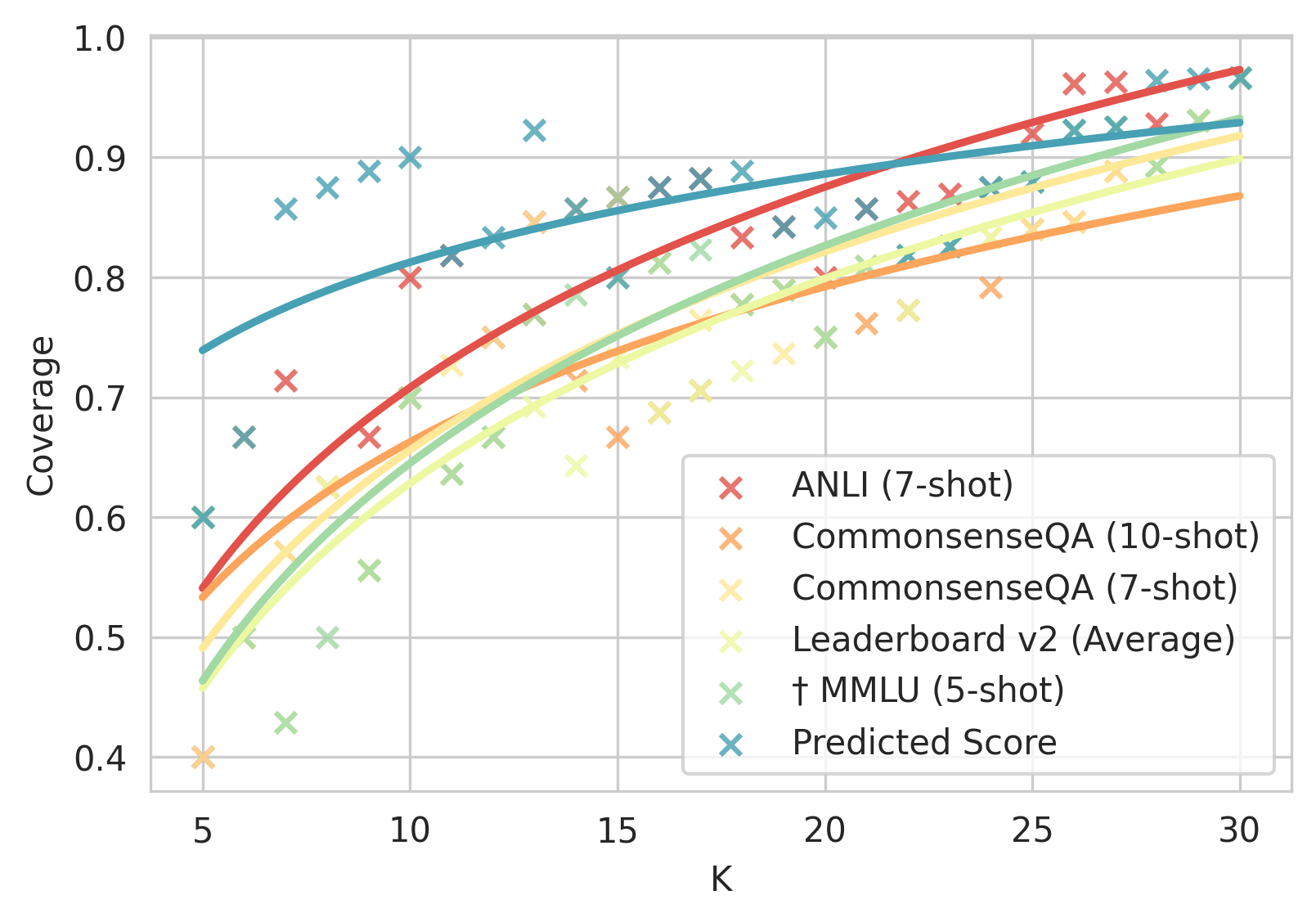}
  \caption{\textbf{Benchmarks vs. Predicted Score Coverage.} We only retain benchmarks with at least 0.4 and 0.7 coverage at $k=5$ and $k=10$, respectively.}
  \label{fig:top_cov_pred}
\end{figure}

\paragraph{Results.}
\autoref{fig:elast_coeff} illustrates the benchmarks with a non-zero weight in the final model.
Mapping back these five benchmarks to their main topics (see \autoref{app:data} for more details), we observe that they consist of two coding (MBPP+~\cite{liu2023is} and HumanEval+~\cite{liu2023is}), one safety (ToxiGen~\cite{hartvigsen-etal-2022-toxigen}), and one general (IFEval~\cite{zhou2023instruction}) benchmarks, along with one training metric (\#Params).
This combination closely follows the subcategories in RewarcBench: Coding $\approx$ Reasoning, Safety $=$ Safety, General + Training Metric $\approx$ Chat/Chat Hard.
Moreover, in \autoref{fig:top_cov_pred}, we compare the coverages of the fitted model to the standalone benchmarks.
As evident, the trained model significantly improves the coverage in lower $K$s, mitigating the critical problem of using standalone benchmarks.
These results prove our hypothesis, showcasing the predictability of reward modeling performance from a low-dimensional vector of prior results.

\begin{table*}
  \centering
  \resizebox{\textwidth}{!}{
  \begin{tabular}{l|cc|cc|cc|cc|cc}
    \toprule

    Model & Chat & $\Delta$ & Chat Hard & $\Delta$ & Safety & $\Delta$ & Reasoning & $\Delta$ & Score & $\Delta$ \\ \midrule

    \texttt{Llama-3.1-8B} & 93.9 & \cellcolor{gray!10}- & 53.7 & \cellcolor{gray!10}- & 64.7 & \cellcolor{gray!10}- & 79.1 & \cellcolor{gray!10}- & 72.9 & \cellcolor{gray!10}- \\ \midrule
    \texttt{Llama-3.1-8B-Instruct} & 95.3 & \cellcolor{PineGreen!10}1.5\% & 68.2 & \cellcolor{PineGreen!35}27.0\% & 84.6 & \cellcolor{PineGreen!39}30.8\% & 84.7 & \cellcolor{PineGreen!15}7.1\% & 83.2 & \cellcolor{PineGreen!22}14.1\% \\ \midrule
    \texttt{Hermes-3-Llama-3.1-8B} & 95.5 &  \cellcolor{PineGreen!10}1.7\% & 71.3 & \cellcolor{PineGreen!41}32.8\% & 83.8 & \cellcolor{PineGreen!38}29.5\% & 74.0 & \cellcolor{red!14}-6.4\% & 81.1 & \cellcolor{PineGreen!19}11.2\% \\ \midrule
    \texttt{Llama-3.1-Tulu-3-8B-SFT} & 95.3 & \cellcolor{PineGreen!10}1.5\% & 70.8 & \cellcolor{PineGreen!40}31.8\% & 84.9 & \cellcolor{PineGreen!39}31.2\% & 85.8 & \cellcolor{PineGreen!17}8.5\% & 84.2 & \cellcolor{PineGreen!24}15.5\% \\
    \texttt{Llama-3.1-Tulu-3-8B-DPO} & 94.7 & \cellcolor{PineGreen!9}0.9\% & 69.1 & \cellcolor{PineGreen!37}28.7\% & 82.3 & \cellcolor{PineGreen!35}27.2\% & 80.1 & \cellcolor{PineGreen!9}1.3\% & 81.6 & \cellcolor{PineGreen!19}11.9\% \\
    \texttt{Llama-3.1-Tulu-3-8B} & 93.3 & \cellcolor{red!9}-0.6\% & 65.6 & \cellcolor{PineGreen!30}22.2\% & 83.5 & \cellcolor{PineGreen!37}29.1\% & 78.5 & \cellcolor{red!9}-0.8\% & 80.2 & \cellcolor{PineGreen!19}10.0\% \\
        
    \bottomrule
  \end{tabular}
  }
  \caption{\textbf{Post-training Performances.} The $\Delta$ columns showcase the relative change to the base model's performance for each category.}
  \label{tab:post}
\end{table*}

\section{Training Stages}
\label{app:comp}

\subsection{Post-training}
\paragraph{Setup.}
Traditionally, for training RMs, practitioners have used a base model that has undergone an SFT process~\cite{stiennon2020learning}.
However, the recent advancements in LLMs have introduced more stages to the training process.
In this section, we analyze the effect of these different stages on the RMs' performance using the publicly available models.
While publishers don't regularly release the intermediate training checkpoints, recent efforts in open LLMs have made some of these intermediate models available for analysis.
Specifically, for the \texttt{Llama-3.1-Tulu-3-8B}\footnote{\href{https://huggingface.co/allenai/Llama-3.1-Tulu-3-8B}{allenai/Llama-3.1-Tulu-3-8B}} model, \citet{lambert2024tulu3} have released three models from the end of each SFT, Direct Preference Optimization (DPO)~\cite{rafailov2023direct}, and Reinforcement Learning with Verifiable Rewards (RLVR) stages.
Apart from the Tulu 3 model, we also include two other \texttt{Llama-3.1-8B}-based\footnote{\href{https://huggingface.co/meta-llama/Llama-3.1-8B}{meta-llama/Llama-3.1-8B}} models that have undergone the post-training phase, namely: \texttt{Llama-3.1-8B-Instruct}\footnote{\href{https://huggingface.co/meta-llama/Llama-3.1-8B-Instruct}{meta-llama/Llama-3.1-8B-Instruct}} and \texttt{Hermes-3-Llama-3.1-8B}\footnote{\href{https://huggingface.co/NousResearch/Hermes-3-Llama-3.1-8B}{NousResearch/Hermes-3-Llama-3.1-8B}; SFT + DPO.}~\cite{teknium2024hermes}.

\paragraph{Results.}
\autoref{tab:post} presents our experimental results comparing different post-training stages to the base model.
From these results, we can observe that the post-training procedure significantly improves the overall performance of RMs.
However, the extra steps after the SFT phase decrease the models' performance across all categories.
This phenomenon could be due to the focus of these stages on human alignment, which slightly degrades other capabilities~\cite{korbak-etal-2022-rl}.
Looking at the subcategories, we note that the \textit{Chat Hard} and \textit{Safety} consistently get significant performance boosts (between 22-32\%) after the post-training procedure.
We believe this is due to dense exposure to related samples that focus on improving the models' safety and complex conversational capabilities.
Moreover, the performances on \textit{Chat} category remain primarily unchanged ($<$2\%), persistent with our previous observations in \autoref{sec:bench} where even the worst models posted high performances.
Finally, in the \textit{Reasoning} category, while the initial SFT stage moderately ($\sim$8.5\%) improves the performance, the following stages reverse most of the gains.
Given the focus of the RLVR stage on improving math capabilities, these results are somewhat surprising.
This phenomenon might be explained by the fact that only 31\% of reasoning samples in RewardBench are math-related, compared to 69\% targeting coding correctness.
However, given a potential co-dependence of math and coding capabilities, further investigation is needed on this phenomenon, which we leave to future works.

\begin{figure*}[t]
  \includegraphics[width=\textwidth]{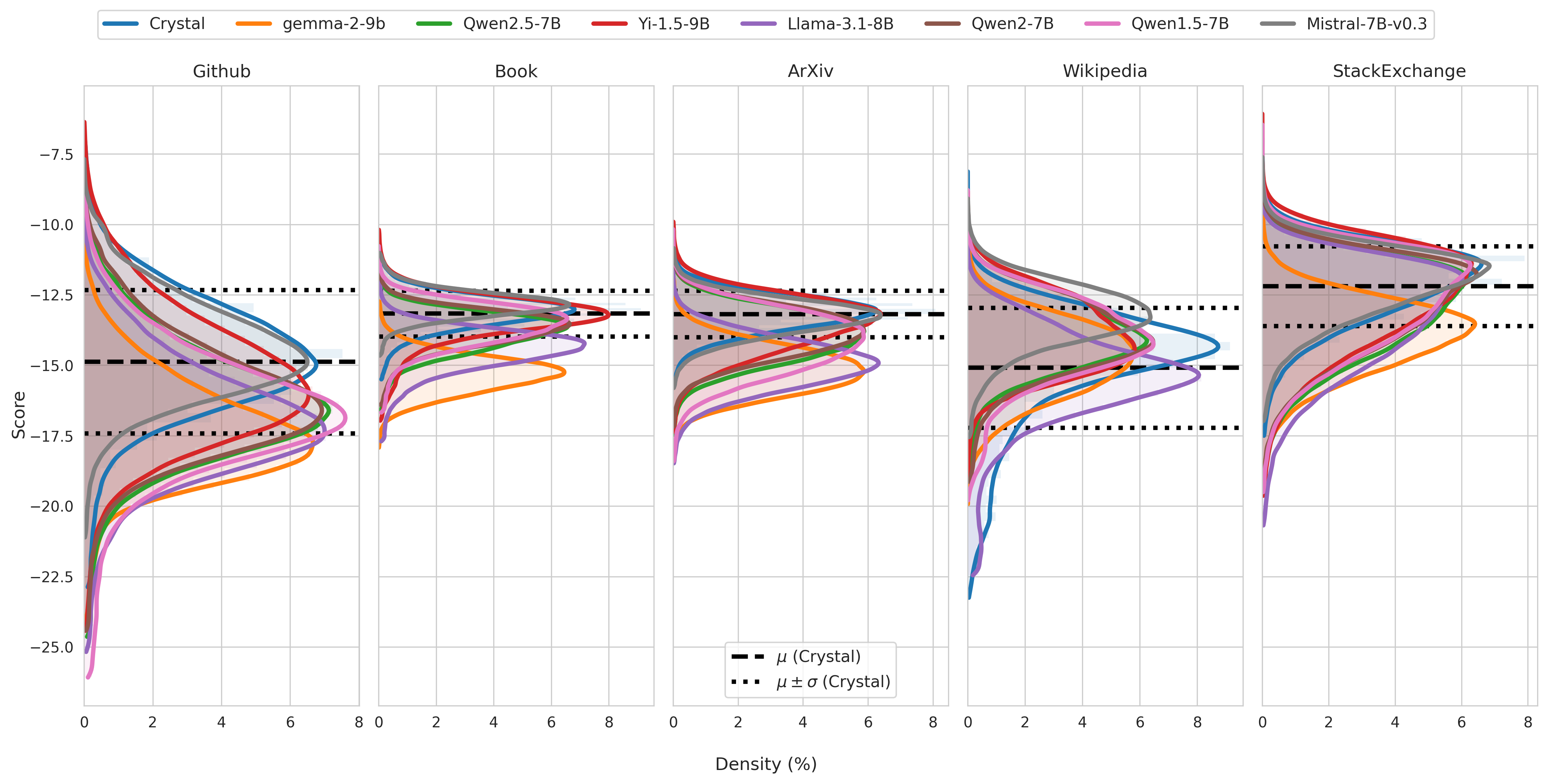}
  \caption{\textbf{Estimated Pre-training Data Distributions.} \texttt{Crystal}~\cite{liu2024llm} represents our ground truth, as it has seen the entire SlimPajama dataset in the pre-training phase exactly once.}
  \label{fig:slimpajama}
\end{figure*}

\begin{figure}[t]
  \includegraphics[width=\columnwidth]{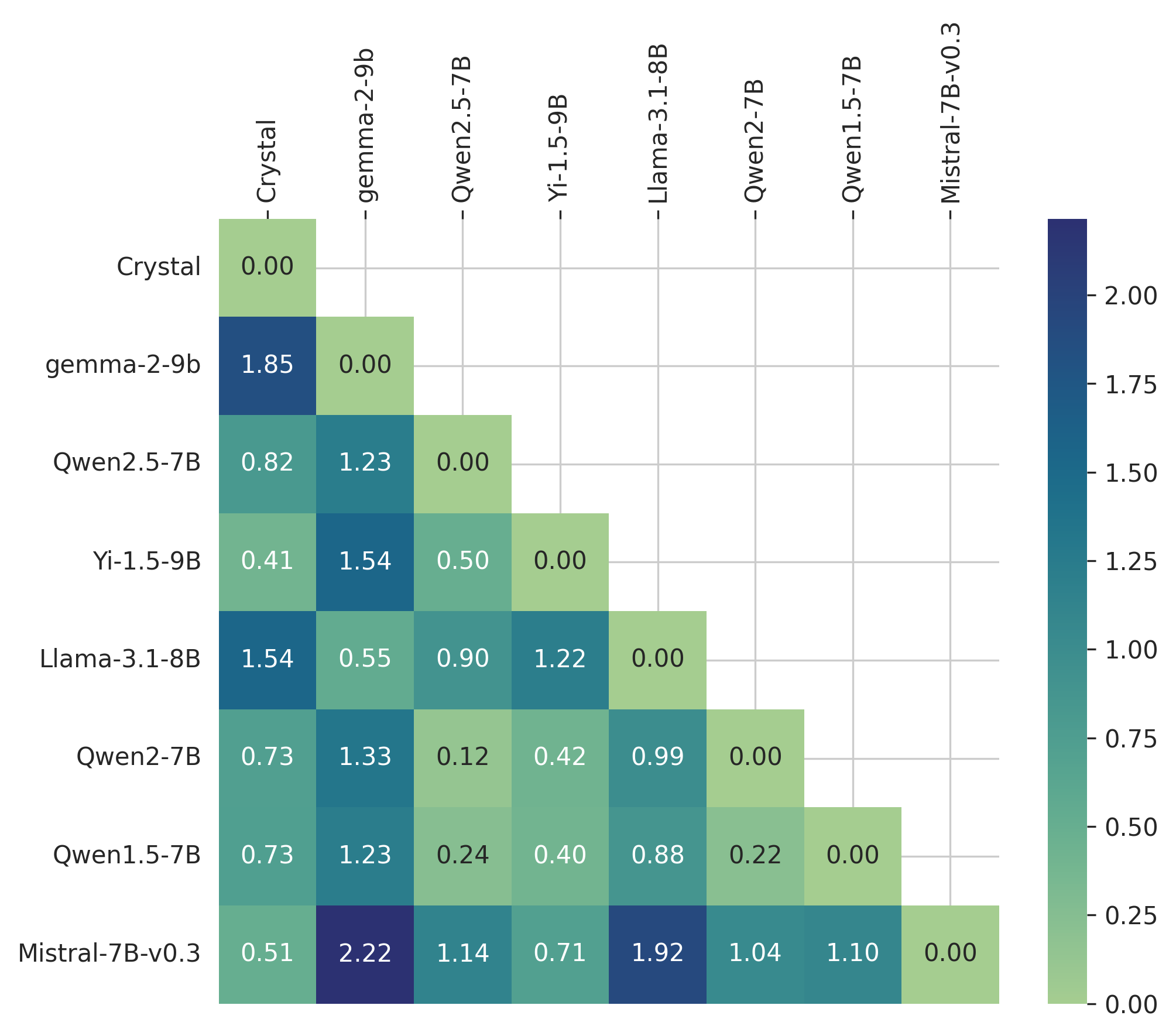}
  \caption{\textbf{Jensen-Shannon Distance.} The values are based on the scores from the entire dataset.}
  \label{fig:slimpajama_jsd}
\end{figure}

\subsection{Pre-training}
\paragraph{Setup.}
Prior works have examined the relation between eventual model capabilities and many LLMs' attributes, ranging from compute~\cite{NEURIPS2022_c1e2faff} to downstream~\cite{ruan2024observational} metrics.
However, pre-training data distribution has remained a significant underexplored factor among these attributes, mainly due to its confidential, proprietary nature.
Efforts in open LLM training~\cite{liu2024llm,olmo20242} present an opportunity to study this factor.
Recent studies~\cite{shi2024detecting,zhang2024pretraining,zhang2024adaptive,kim2024detecting} have developed pre-training data detection techniques by viewing it as a membership inference attack (MIA) task.
However, the curated MIA datasets lack the scale and coverage needed for a comprehensive analysis of the pre-training data distribution, as they have less than 10k samples.
To address this issue, we curate a large-scale dataset by sampling 200k examples from each of the \textit{Github}, \textit{Book}, \textit{ArXiv}, \textit{Wikipedia}, and \textit{StackExchange} subsets in SlimPajama~\cite{soboleva2023slimpajama}, resulting in a 1M sample dataset\footnote{1.25\% of all the documents in the original categories.}.
Moreover, to detect the presence of a document in an LLM, we use a truncated version (\textit{i.e.,} the first 2048 tokens) of the length-normalized sequence probability~\cite{malinin2021uncertainty}.
The truncation helps reduce the cost of running such analysis at scale, as some books have more than 170k tokens, and mitigates the noise from later tokens, as LLMs have shown to have a problem making robust use of tokens in the middle of long documents~\cite{liu-etal-2024-lost,hsieh2024ruler}.

Given a document $D = [t_i]_{i=1,\dots,m}$, an LLM $\phi$, and a tokens limit $N$, we calculate a presence score $\mathcal{S}_\phi$ as
\begin{equation}
   \mathcal{S}_\phi(D, N) = \frac{1}{N}\sum_{i=1}^N \log p_\phi(t_i | t_{1:i-1}) \  .
\end{equation}
We use \texttt{Crystal}\footnote{\href{https://huggingface.co/LLM360/Crystal}{LLM360/Crystal}}~\cite{liu2024llm} as our ground truth LLM, as all of the SlimPajama dataset has been used in its pre-training stage.
Finally, for each model, we reuse the extracted distribution from the largest member of its family if and only if they've been trained on the same amount of data, assuming the same data was used for the pre-training stage (see \autoref{app:data} for more details).

\paragraph{Results.}
\autoref{fig:slimpajama} illustrates the score distributions across different subsets of SlimPajama for seven models from different families.
Notably, we observe a difference between the score ranges across the categories, even for the ground truth model that has seen everything once.
We believe this is due to the potential occurrence of similar documents in the excluded \textit{CommonCrawl} and \textit{C4} categories.
\autoref{fig:slimpajama} showcases the Jensen-Shannon Distance (JSD) between different models over the scores of the entire 1M samples.
As evident, some model pairs showcase significantly higher distances than others, showcasing a variability across models that can be utilized for downstream predictions.
We also notice that the Qwen$\{1.5,2,2.5\}$ models have the lowest non-zero distances, which suggests that different generations of models released by a publisher potentially have significant overlaps in their pre-training data.
Moreover, we expand our regression analysis (see \autoref{sec:bench:reg}) by adding the average scores of the categories to the already established five features (see \autoref{fig:elast_coeff}).
Our experiments show that compared to adding these features improves the mean absolute error by $+$1.5\% (from 3.2\% to 1.7\%), compared to only using the original five features, which showcases the untapped potential of the pre-training data distributions.

\section{Conclusion}
In this paper, we presented a systematic analysis of the effect of base model selection on the reward modeling performance.
First, we showcased the significant variability of final performance by only changing the base model.
Then, we analyzed the possibility of knowing a model's performance apriori, leading to a simple model with high coverage across the range of models, using commonly disclosed metrics and performances.
Finally, we investigate different training stages, showcasing 1) the positive and negative effects of certain steps in post-training and 2) illustrating the untapped potential of using estimated pre-training data distributions.

\section*{Limitations}
\paragraph{Training Regimen.}
While our experiments are designed to remove the effect of reward modeling training data (\textit{i.e.,} using the same small dataset for all models), using larger datasets might reveal unknown behaviors for some models.
However, given our computational resource constraints, we leave these experiments to future works, as the current cost of our experiments is $\sim$4500 GPU/hours.

\paragraph{Post-training.}
In our analysis, we observed an interesting and unintuitive phenomenon where RLHF and preference optimization hurt the models' performance in the reasoning category of RewardBench.
However, we only had access to a limited number of publicly available models; further investigation is needed to exhibit the main reason for this phenomenon.

\paragraph{Pre-training.}
Given our limited resources, we could only run our data distribution estimation experiments on a subset of models.
Extending our model set in future works will boost our understanding of the effect of data distributions.
Moreover, we relied on a relatively simple score to scale to the number of samples we had; further experiments with other methods at scale could help gain more insights.

\section*{Acknowledgments}
The authors thank Alon Benhaim, Barun Patra, Xihui Lin, and Dong-Ho Lee for insightful discussions.
This work was funded by the Defense Advanced Research Projects Agency with award HR00112220046.

\bibliography{main}

\newpage
\appendix

\section{RewardBench as Ground Truth}
Given the heavy reliance of our work on RewardBench, we conduct an independent verification of the preferences.
Specifically, we sample 50 tasks from the tasks that our top 10 models got wrong the most.
Then, we gather 3 annotations from different annotators and use a majority vote to determine the final preference.
All annotators were senior Computer Science PhD students specializing in NLP with extensive experience working with and evaluating LLMs.
Our results show an agreement of 98\%, establishing the quality of RewardBench.

\section{Benchmarks}
\label{app:data}
\begin{figure}[t]
  \includegraphics[width=\columnwidth]{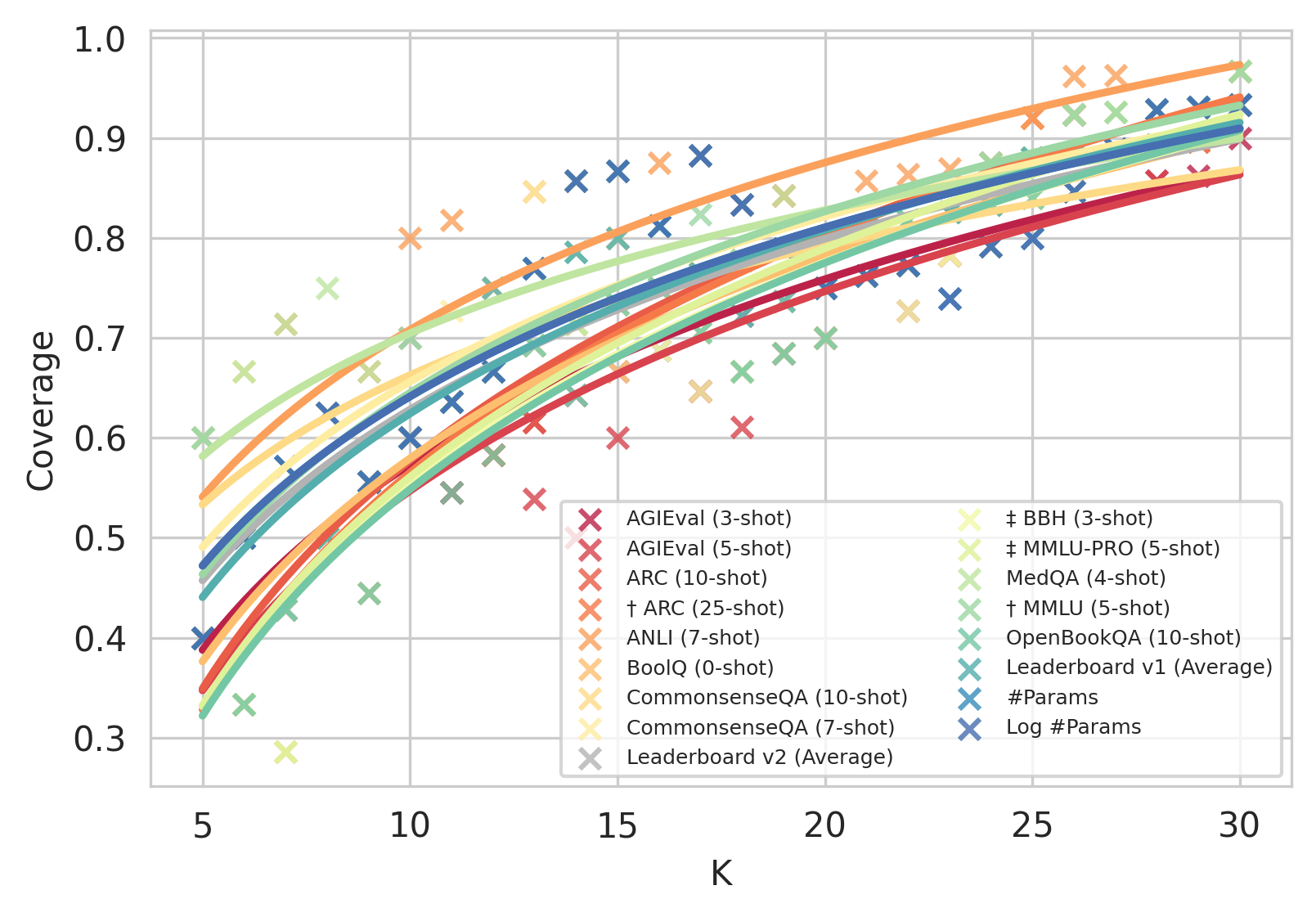}
  \caption{\textbf{Benchmark's Coverage.} We only retain benchmarks with at least 0.4 and 0.6 coverage at $k=5$ and $k=10$, respectively.}
  \label{fig:more_cov}
\end{figure}

\autoref{tab:datasets} showcases all the 32 benchmarks used in our experiments.
Moreover, \autoref{fig:more_cov} illustrates the coverage using an expanded set of benchmarks with at least 0.4 and 0.6 coverage at k = 5 and k = 10, respectively.

\begin{table*}
  \centering
  \resizebox{\textwidth}{!}{
  \begin{tabular}{ccccc}
    \toprule
    \textbf{Framework} & \textbf{Dataset} & \textbf{Topic} & \textbf{\#Shots} & \textbf{Models} \\ \midrule

    \multirow{32}{*}{\makecell{\texttt{lm\_eval}\\\cite{eval-harness}}}
    & \texttt{leaderboard\_ifeval}~\cite{zhou2023instruction} &  \multirow{2}{*}{General} & 0 & \textcolor{blue}{L}\textcolor{red}{G}\textcolor{OliveGreen}{P}\textcolor{violet}{Q} \\
    & \texttt{winogrande}~\cite{sakaguchi2021winogrande} & & 5 & \textcolor{blue}{L}\textcolor{red}{G}\textcolor{OliveGreen}{P} \\
    \cmidrule{2-5}

    & \texttt{hellaswag}~\cite{zellers-etal-2019-hellaswag} &  \multirow{6}{*}{Reading Comprehension} & 5,10 & \textcolor{red}{G}\textcolor{OliveGreen}{P} \\
    & \texttt{openbookqa}~\cite{mihaylov-etal-2018-suit} & & 10 & \textcolor{OliveGreen}{P} \\
    & \texttt{triviaqa}~\cite{joshi-etal-2017-triviaqa} & & 5 & \textcolor{blue}{L}\textcolor{red}{G}\textcolor{OliveGreen}{P} \\
    & \texttt{squadv2}~\cite{rajpurkar-etal-2018-know} & & 1 & \textcolor{blue}{L} \\
    & \texttt{drop}~\cite{dua-etal-2019-drop} & & 3 & \textcolor{blue}{L} \\
    & \texttt{boolq}~\cite{clark-etal-2019-boolq} & & 0 & \textcolor{blue}{L}\textcolor{red}{G}\textcolor{OliveGreen}{P} \\
    \cmidrule{2-5}

    & \texttt{anli}~\cite{zhong-etal-2024-agieval} &  \multirow{2}{*}{Adversarial} & 7 & \textcolor{OliveGreen}{P} \\
    & \texttt{truthfulqa\_mc2}~\cite{lin-etal-2022-truthfulqa} & & 10 & \textcolor{red}{G}\textcolor{OliveGreen}{P} \\
    \cmidrule{2-5}

    & \texttt{commonsense\_qa}~\cite{talmor-etal-2019-commonsenseqa} &  \multirow{4}{*}{Commonsense Reasoning} & 7,10 & \textcolor{blue}{L}\textcolor{OliveGreen}{P} \\
    & \texttt{piqa}~\cite{bisk2020piqa} & & 0,5 & \textcolor{red}{G}\textcolor{OliveGreen}{P} \\
    & \texttt{social\_iqa}~\cite{sap-etal-2019-social} & & 0,5 & \textcolor{red}{G}\textcolor{OliveGreen}{P} \\
    & \texttt{nq\_open}~\cite{kwiatkowski-etal-2019-natural} &  & 5 & \textcolor{red}{G} \\  \cmidrule{2-5}
    \cmidrule{2-5}

    & \texttt{agieval\_en}~\cite{zhong-etal-2024-agieval} &  \multirow{8}{*}{Expert Reasoning} & 3,5 & \textcolor{blue}{L}\textcolor{red}{G}\textcolor{OliveGreen}{P} \\
    & \texttt{ai2\_arc}~\cite{clark2018think} & & 0,10,25 & \textcolor{blue}{L}\textcolor{red}{G}\textcolor{OliveGreen}{P} \\
    & \texttt{leaderboard\_bbh}~\cite{suzgun-etal-2023-challenging} & & 3 & \textcolor{blue}{L}\textcolor{red}{G}\textcolor{OliveGreen}{P}\textcolor{violet}{Q} \\
    & \texttt{leaderboard\_gpqa}~\cite{rein2024gpqa} & & 0 & \textcolor{blue}{L}\textcolor{red}{G}\textcolor{OliveGreen}{P}\textcolor{violet}{Q} \\
    & \texttt{leaderboard\_mmlu\_pro}~\cite{wang2024mmlupro} & & 5 & \textcolor{blue}{L}\textcolor{red}{G}\textcolor{OliveGreen}{P}\textcolor{violet}{Q} \\
    & \texttt{leaderboard\_musr}~\cite{gao-etal-2023-learning-multilingual} & & 0 & \textcolor{blue}{L}\textcolor{red}{G}\textcolor{OliveGreen}{P}\textcolor{violet}{Q} \\
    & \texttt{medqa\_4options}~\cite{jin2021disease} & & 2 & \textcolor{OliveGreen}{P} \\
    & \texttt{mmlu}~\cite{hendrycks2021measuring} & & 5 & \textcolor{blue}{L}\textcolor{red}{G}\textcolor{OliveGreen}{P} \\
    \cmidrule{2-5}

    & \texttt{gsm8k\_cot\_llama}~\cite{cobbe2021training} &  \multirow{2}{*}{Math} & 5,8 & \textcolor{blue}{L}\textcolor{red}{G}\textcolor{OliveGreen}{P}\textcolor{violet}{Q} \\
    & \texttt{leaderboard\_math}~\cite{hendrycks2021measuring} & & 4 & \textcolor{blue}{L}\textcolor{red}{G}\textcolor{OliveGreen}{P}\textcolor{violet}{Q} \\
    \cmidrule{2-5}

    & \texttt{crows\_pairs\_english}~\cite{nangia-etal-2020-crows} &  \multirow{2}{*}{Safety} & 0 & \textcolor{red}{G} \\
    & \texttt{toxigen}~\cite{hartvigsen-etal-2022-toxigen} & & 0 & \textcolor{red}{G} \\ 
    \cmidrule{2-5}
    & \texttt{qasper}~\cite{dasigi-etal-2021-dataset} & Long-context & 0 & \textcolor{OliveGreen}{P} \\
    \cmidrule{2-5}
    & \texttt{leaderboard} v1~\cite{open-llm-leaderboard-v1} &  \multirow{2}{*}{Aggregate} & - & \textcolor{blue}{L}\textcolor{red}{G}\textcolor{OliveGreen}{P}\textcolor{violet}{Q} \\
    & \texttt{leaderboard} v2~\cite{open-llm-leaderboard-v2} & & - & \textcolor{blue}{L}\textcolor{red}{G}\textcolor{OliveGreen}{P}\textcolor{violet}{Q} \\
    \midrule

    \multirow{4}{*}{\makecell{\texttt{evalplus}\\\cite{liu2023is}}}
    & \texttt{HumanEval}~\cite{chen2021codex} & \multirow{4}{*}{Coding} & 0 & \textcolor{blue}{L}\textcolor{red}{G}\textcolor{OliveGreen}{P}\textcolor{violet}{Q} \\
    & \texttt{HumanEval+}~\cite{liu2023is} & & 0 & \textcolor{blue}{L}\textcolor{red}{G}\textcolor{OliveGreen}{P}\textcolor{violet}{Q} \\
    & \texttt{MBPP}~\cite{austin2021program} & & 0 & \textcolor{blue}{L}\textcolor{red}{G}\textcolor{OliveGreen}{P}\textcolor{violet}{Q} \\
    & \texttt{MBPP+}~\cite{liu2023is} & & 0 & \textcolor{blue}{L}\textcolor{red}{G}\textcolor{OliveGreen}{P}\textcolor{violet}{Q} \\

    \bottomrule
  \end{tabular}
  }
  \caption{\textbf{Benchmarks.} We gather a comprehensive list of 33 common benchmarks from the technical reports of well-known models. \textbf{Legened:} \textcolor{blue}{L} $\rightarrow$ Llama-3.x, \textcolor{red}{G} $\rightarrow$ Gemma-2, \textcolor{OliveGreen}{P} $\rightarrow$ Phi-3.5, and \textcolor{violet}{Q} $\rightarrow$ Qwen2.5.}
  \label{tab:datasets}
\end{table*}

\section{Models}
\label{app:model}
\autoref{tab:models} showcases all the 40 models used in our experiments.

\begin{table*}
  \centering
  \resizebox{\textwidth}{!}{
  \begin{tabular}{ccccccc}
    \toprule
    \textbf{Publisher} & \textbf{Model} & \textbf{\makecell{Release Date\\(First Commit)}} & \textbf{\makecell{\#Params\\(B)}} & \textbf{\makecell{\#Downloads\\(Feb
    2025)}} & \textbf{\#Likes} &\textbf{\makecell{\#Pre-training Tokens\\(T)}} \\ \midrule

    \multirow{3}{*}{Microsoft}
    & \texttt{Phi-3.5-mini-instruct} & 08/2024 & 3.82 & 1.143M & 776 &3.4 \\ \cmidrule{2-7}
    & \texttt{Phi-3-small-8k-instruct} & 05/2024 & 7.38 & 25.1k&160 & 4.8 \\
    & \texttt{Phi-3-mini-4k-instruct} & 04/2024 & 3.82 & 900k &1122 &3.3 \\ \midrule

    \multirow{6}{*}{Google} 
    & \texttt{gemma-2-9b-it} & 06/2024 & 9.24 & 393.4k&639& 8.0 \\
    & \texttt{gemma-2-2b-it} & 07/2024 & 2.61 & 437.6k&915 & 2.0 \\ \cmidrule{2-7}
    & \texttt{gemma-1.1-7b-it} & 03/2024 & 8.54 & 20.7k & 270&6.0 \\
    & \texttt{gemma-1.1-2b-it} & 03/2024 & 2.51 & 93.3k& 154& 6.0 \\ \cmidrule{2-7}
    & \texttt{gemma-7b-it} & 02/2024 & 8.54 & 62.1k & 1151&6.0 \\
    & \texttt{gemma-2b-it} & 02/2024 & 2.51 & 105.8k & 701&6.0 \\ \midrule

    \multirow{4}{*}{Meta}
    & \texttt{Llama-3.2-3B-Instruct} & 09/2024 & 3.21 & 1.497M & 939& 9.0 \\
    & \texttt{Llama-3.2-1B-Instruct} & 09/2024 & 1.24 & 1.523M &738& 9.0 \\ \cmidrule{2-7}
    & \texttt{Llama-3.1-8B-Instruct} & 07/2024 & 8.03 & 5.669M & 3546 & 15.0 \\ \cmidrule{2-7}
    & \texttt{Meta-Llama-3-8B-Instruct} & 04/2024 & 8.03 & 2.101M & 3788& 15.0 \\ \midrule

    \multirow{3}{*}{01.ai} 
    & \texttt{Yi-1.5-9B-Chat} & 05/2024 & 8.83 & 20.9k&139 & 3.6 \\
    & \texttt{Yi-1.5-6B-Chat} & 05/2024 & 6.06 & 19.6k
&41 & 3.6 \\ \cmidrule{2-7}
    & \texttt{Yi-6B-Chat} & 11/2023 & 6.06 & 9.3k &65& 3.0 \\ \midrule
    
    \multirow{11}{*}{Alibaba}
    & \texttt{Qwen2.5-7B-Instruct} & 09/2024 & 7.62 & 1.275M&459 & 18.0 \\
    & \texttt{Qwen2.5-3B-Instruct} & 09/2024 & 3.09 & 326.5k & 158&18.0 \\
    & \texttt{Qwen2.5-1.5B-Instruct} & 09/2024 & 1.54 & 592.5k&299&  18.0 \\
    & \texttt{Qwen2.5-0.5B-Instruct} & 09/2024 & 0.49 & 696.2k&198 & 18.0 \\ \cmidrule{2-7}
    
    & \texttt{Qwen2-7B-Instruct} & 06/2024 & 7.62 & 821.4k & 611 &7.0 \\
    & \texttt{Qwen2-1.5B-Instruct} & 06/2024 & 1.54 & 187.9k&134 & 7.0 \\
    & \texttt{Qwen2-0.5B-Instruct} & 06/2024 & 0.49 & 170.3k&174 & 12.0 \\ \cmidrule{2-7}
    
    & \texttt{Qwen1.5-7B-Chat} & 01/2024 & 7.72 & 25.5k&165 & 4.0 \\
    & \texttt{Qwen1.5-4B-Chat} & 01/2024 & 3.95 & 5.6k&38 & 2.4 \\
    & \texttt{Qwen1.5-1.8B-Chat} & 01/2024 & 1.84 & 11.2k & 48&2.4 \\
    & \texttt{Qwen1.5-0.5B-Chat} & 01/2024 & 0.62 & 556.2k &76 & 2.4 \\ \midrule

    \multirow{3}{*}{Mistral AI}
    & \texttt{Mistral-7B-Instruct-v0.3} & 05/2024 & 7.25 & 1.755M & 1293 & 8.0 \\
    & \texttt{Mistral-7B-Instruct-v0.2} & 12/2023 & 7.24 & 3.586M& 2634 & 8.0 \\
    & \texttt{Mistral-7B-Instruct-v0.1} & 09/2023 & 7.24 & 1.332M & 1547 &8.0 \\ \midrule

    \multirow{1}{*}{Stability AI}
    & \texttt{stablelm-2-1\_6b-chat} & 04/2024 & 1.64 & 4.4k&32 & 2.0 \\ \midrule

    \multirow{2}{*}{Nvidia}
    & \texttt{Mistral-NeMo-Minitron-8B-Instruct} & 10/2024 & 8.41 & 3.1k&71 & 15.0 \\
    & \texttt{Nemotron-Mini-4B-Instruct} & 09/2024 & 4.20 & 0.1k&147 & 8.0 \\ \midrule

    \multirow{3}{*}{Ai2}
    & \texttt{Llama-3.1-Tulu-3-8B-SFT} & 11/2024 & 8.03 & 23.4k&21& 15.0 \\
    & \texttt{Llama-3.1-Tulu-3-8B-DPO} & 11/2024 & 8.03 & 28.5k&22 & 15.0 \\
    & \texttt{Llama-3.1-Tulu-3-8B} & 11/2024 & 8.03 & 12.7k &139& 15.0 \\ \midrule

    \multirow{4}{*}{TII}
    & \texttt{Falcon3-10B-Instruct} & 12/2024 & 10.30 & 37,9k&87 & 16.0 \\
    & \texttt{Falcon3-7B-Instruct} & 12/2024 & 7.46 & 45.2k&49 & 14.0 \\
    & \texttt{Falcon3-3B-Instruct} & 12/2024 & 3.23 & 30.5k&23 & 14.1 \\
    & \texttt{Falcon3-1B-Instruct} & 12/2024 & 1.67 & 31.4k & 32&14.1 \\
    \bottomrule
  \end{tabular}
  }
  \caption{\textbf{Models.} We curate an inclusive list of 40 models from prominent model providers.}
  \label{tab:models}
\end{table*}

\section{Full Results}
\label{app:full}
\autoref{tab:bt_perf} and \autoref{tab:reg_perf} present the full results using the Bradley-Terry and Regression methods, respectively.

\begin{table*}
  \centering
  \resizebox{\textwidth}{!}{
    \begin{tabular}{ccccccc}
    \toprule
    \textbf{Publisher} & \textbf{Model} & \textbf{Chat} & \textbf{Chat Hard} & \textbf{Safety} & \textbf{Reasoning} & \textbf{Score} \\
    \midrule
    \multirow{3}{*}{Microsoft}
    & \texttt{Phi-3.5-mini-instruct} & 96.1 & 62.3 & 77.2 & 76.9 & 78.1 \\
    \cmidrule{2-7}
    & \texttt{Phi-3-small-8k-instruct} & 89.7 & 66.7 & 76.4 & 57.0 & 72.4 \\
    & \texttt{Phi-3-mini-4k-instruct} & 96.4 & 58.6 & 77.2 & 83.6 & 78.9 \\
    \midrule

    \multirow{6}{*}{Google}
    & \texttt{gemma-2-9b-it} & 95.8 & 74.1 & 88.4 & 94.3 & 88.1 \\
    & \texttt{gemma-2-2b-it} & 94.7 & 56.8 & 79.9 & 80.7 & 78.0 \\
    \cmidrule{2-7}
    & \texttt{gemma-1.1-7b-it} & 97.2 & 61.0 & 81.1 & 79.5 & 79.7 \\
    & \texttt{gemma-1.1-2b-it} & 89.4 & 46.3 & 74.6 & 50.5 & 65.2 \\
    \cmidrule{2-7}
    & \texttt{gemma-7b-it} & 93.3 & 60.5 & 83.4 & 78.1 & 78.8 \\
    & \texttt{gemma-2b-it} & 92.2 & 42.5 & 67.0 & 56.7 & 64.6 \\ \midrule

    \multirow{4}{*}{Meta}
    & \texttt{Llama-3.2-3B-Instruct} & 95.3 & 68.6 & 87.7 & 59.3 & 77.7 \\
    & \texttt{Llama-3.2-1B-Instruct} & 93.3 & 42.3 & 65.4 & 70.2 & 67.8 \\
    \cmidrule{2-7}
    & \texttt{Llama-3.1-8B-Instruct} & 95.3 & 68.2 & 84.6 & 84.7 & 83.2 \\
    \cmidrule{2-7}
    & \texttt{Meta-Llama-3-8B-Instruct} & 93.9 & 75.4 & 86.6 & 81.2 & 84.3 \\
    \midrule

    \multirow{3}{*}{01.AI}
    & \texttt{Yi-1.5-9B-Chat} & 95.8 & 69.5 & 80.1 & 88.7 & 83.5 \\
    & \texttt{Yi-1.5-6B-Chat} & 93.3 & 63.4 & 77.2 & 78.3 & 78.0 \\
    \cmidrule{2-7}
    & \texttt{Yi-6B-Chat} & 93.3 & 56.4 & 71.5 & 67.4 & 72.2 \\
    \midrule

    \multirow{11}{*}{Alibaba}
    & \texttt{Qwen2.5-7B-Instruct} & 94.7 & 72.8 & 87.8 & 90.7 & 86.5 \\
    & \texttt{Qwen2.5-3B-Instruct} & 92.7 & 63.4 & 82.0 & 85.3 & 80.8 \\
    & \texttt{Qwen2.5-1.5B-Instruct} & 92.7 & 56.4 & 80.7 & 84.8 & 78.6 \\
    & \texttt{Qwen2.5-0.5B-Instruct} & 89.9 & 45.6 & 51.9 & 48.4 & 59.0 \\
    \cmidrule{2-7}
    & \texttt{Qwen2-7B-Instruct} & 95.3 & 66.4 & 78.4 & 84.0 & 81.0 \\
    & \texttt{Qwen2-1.5B-Instruct} & 92.7 & 47.8 & 72.0 & 79.0 & 72.9 \\
    & \texttt{Qwen2-0.5B-Instruct} & 92.2 & 39.9 & 54.7 & 60.7 & 61.9 \\
    \cmidrule{2-7}
    & \texttt{Qwen1.5-7B-Chat} & 93.3 & 51.8 & 74.6 & 81.3 & 75.2 \\
    & \texttt{Qwen1.5-4B-Chat} & 91.1 & 50.9 & 78.0 & 77.6 & 74.4 \\
    & \texttt{Qwen1.5-1.8B-Chat} & 90.8 & 40.1 & 56.4 & 64.8 & 63.0 \\
    & \texttt{Qwen1.5-0.5B-Chat} & 91.3 & 43.2 & 58.0 & 58.0 & 62.6 \\
    \midrule

    \multirow{3}{*}{Mistral AI}
    & \texttt{Mistral-7B-Instruct-v0.3} & 94.1 & 62.3 & 75.1 & 84.1 & 78.9 \\
    & \texttt{Mistral-7B-Instruct-v0.2} & 93.0 & 59.9 & 78.2 & 79.5 & 77.6 \\
    & \texttt{Mistral-7B-Instruct-v0.1} & 92.7 & 58.8 & 71.1 & 71.8 & 73.6 \\
    \midrule

    \multirow{1}{*}{Stability AI}
    & \texttt{stablelm-2-1\_6b-chat} & 90.5 & 47.4 & 59.3 & 69.0 & 66.5 \\
    \midrule

    \multirow{2}{*}{Nvidia}
    & \texttt{Mistral-NeMo-Minitron-8B-Instruct} & 93.6 & 61.0 & 82.6 & 82.9 & 80.0 \\
    & \texttt{Nemotron-Mini-4B-Instruct} & 93.0 & 61.4 & 75.0 & 82.0 & 77.8 \\
     \midrule

    \multirow{3}{*}{Ai2}
    & \texttt{Llama-3.1-Tulu-3-8B-SFT} & 95.3 & 70.8 & 84.9 & 85.8 & 84.2 \\
    & \texttt{Llama-3.1-Tulu-3-8B-DPO} & 94.7 & 69.1 & 82.3 & 80.1 & 81.6 \\
    & \texttt{Llama-3.1-Tulu-3-8B} & 93.3 & 65.6 & 83.5 & 78.5 & 80.2 \\
    \midrule

    \multirow{4}{*}{TII}
    & \texttt{Falcon3-7B-Instruct} & 96.6 & 64.0 & 89.7 & 80.4 & 82.7 \\
    & \texttt{Falcon3-3B-Instruct} & 95.0 & 53.9 & 78.1 & 73.9 & 75.2 \\
    & \texttt{Falcon3-1B-Instruct} & 84.6 & 31.6 & 53.2 & 46.2 & 53.9 \\
    & \texttt{Falcon3-10B-Instruct} & 95.5 & 67.3 & 89.5 & 91.1 & 85.9 \\
    \bottomrule
    \end{tabular}
  }
  \caption{\textbf{Regression Performance.}}
  \label{tab:reg_perf}
\end{table*}

\begin{table*}
  \centering
  \resizebox{\textwidth}{!}{
    \begin{tabular}{ccccccc}
    \toprule
    \textbf{Publisher} & \textbf{Model} & \textbf{Chat} & \textbf{Chat Hard} & \textbf{Safety} & \textbf{Reasoning} & \textbf{Score} \\
    \midrule
    \multirow{3}{*}{Microsoft}
    & \texttt{Phi-3.5-mini-instruct} & 61.5 & 51.5 & 63.1 & 61.1 & 59.3 \\
    \cmidrule{2-7}
    & \texttt{Phi-3-small-8k-instruct} & 83.5 & 55.3 & 81.9 & 75.8 & 74.1 \\
    & \texttt{Phi-3-mini-4k-instruct} & 64.8 & 46.1 & 56.6 & 59.7 & 56.8 \\
    \midrule

    \multirow{6}{*}{Google}
    & \texttt{gemma-2-9b-it} & 83.8 & 51.1 & 70.8 & 83.6 & 72.3 \\
    & \texttt{gemma-2-2b-it} & 84.1 & 46.5 & 67.6 & 81.3 & 69.9 \\
    \cmidrule{2-7}
    & \texttt{gemma-1.1-7b-it} & 76.3 & 45.4 & 65.4 & 75.1 & 65.5 \\
    & \texttt{gemma-1.1-2b-it} & 74.0 & 41.9 & 67.0 & 63.2 & 61.5 \\
    \cmidrule{2-7}
    & \texttt{gemma-7b-it} & 77.1 & 43.0 & 63.8 & 72.5 & 64.1 \\
    & \texttt{gemma-2b-it} & 79.6 & 39.0 & 65.0 & 63.7 & 61.8 \\
    \midrule

    \multirow{4}{*}{Meta}
    & \texttt{Llama-3.2-3B-Instruct} & 70.4 & 47.4 & 50.8 & 58.9 & 56.9 \\
    & \texttt{Llama-3.2-1B-Instruct} & 57.0 & 51.3 & 58.0 & 56.0 & 55.6 \\
    \cmidrule{2-7}
    & \texttt{Llama-3.1-8B-Instruct} & 78.2 & 62.1 & 69.5 & 65.1 & 68.7 \\
    \cmidrule{2-7}
    & \texttt{Meta-Llama-3-8B-Instruct} & 73.2 & 53.9 & 57.2 & 59.1 & 60.9 \\
    \midrule

    \multirow{3}{*}{01.AI}
    & \texttt{Yi-1.5-9B-Chat} & 80.7 & 54.8 & 62.8 & 67.4 & 66.4 \\
    & \texttt{Yi-1.5-6B-Chat} & 76.5 & 50.2 & 59.9 & 81.3 & 67.0 \\
    \cmidrule{2-7}
    & \texttt{Yi-6B-Chat} & 71.5 & 52.9 & 67.0 & 71.6 & 65.7 \\
    \midrule

    \multirow{11}{*}{Alibaba}
    & \texttt{Qwen2.5-7B-Instruct} & 90.5 & 61.8 & 78.1 & 74.1 & 76.1 \\
    & \texttt{Qwen2.5-3B-Instruct} & 74.0 & 57.0 & 75.1 & 75.8 & 70.5 \\
    & \texttt{Qwen2.5-1.5B-Instruct} & 80.2 & 49.6 & 58.4 & 78.1 & 66.6 \\
    & \texttt{Qwen2.5-0.5B-Instruct} & 79.1 & 42.5 & 55.3 & 69.5 & 61.6 \\
    \cmidrule{2-7}
    & \texttt{Qwen2-7B-Instruct} & 85.5 & 51.1 & 57.8 & 76.7 & 67.8 \\
    & \texttt{Qwen2-1.5B-Instruct} & 70.7 & 47.4 & 56.1 & 69.0 & 60.8 \\
    & \texttt{Qwen2-0.5B-Instruct} & 70.4 & 48.0 & 57.0 & 67.8 & 60.8 \\
    \cmidrule{2-7}
    & \texttt{Qwen1.5-7B-Chat} & 77.7 & 51.3 & 62.3 & 69.3 & 65.1 \\
    & \texttt{Qwen1.5-4B-Chat} & 75.4 & 48.9 & 53.0 & 66.6 & 61.0 \\
    & \texttt{Qwen1.5-1.8B-Chat} & 79.9 & 40.4 & 59.9 & 62.9 & 60.8 \\
    & \texttt{Qwen1.5-0.5B-Chat} & 71.5 & 44.1 & 60.3 & 54.7 & 57.7 \\
    \midrule

    \multirow{3}{*}{Mistral AI}
    & \texttt{Mistral-7B-Instruct-v0.3} & 56.7 & 53.1 & 58.2 & 50.0 & 54.5 \\
    & \texttt{Mistral-7B-Instruct-v0.2} & 80.7 & 38.2 & 54.1 & 58.1 & 57.8 \\
    & \texttt{Mistral-7B-Instruct-v0.1} & 56.7 & 52.6 & 58.4 & 57.2 & 56.2 \\
    \midrule

    \multirow{1}{*}{Stability AI}
    & \texttt{stablelm-2-1\_6b-chat} & 71.2 & 49.3 & 60.5 & 59.9 & 60.2 \\
    \midrule

    \multirow{2}{*}{Nvidia}
    & \texttt{Mistral-NeMo-Minitron-8B-Instruct} & 86.3 & 50.2 & 56.9 & 77.4 & 67.7 \\
    & \texttt{Nemotron-Mini-4B-Instruct} & 81.6 & 49.8 & 63.2 & 50.9 & 61.4 \\
    \midrule

    \multirow{3}{*}{Ai2}
    & \texttt{Llama-3.1-Tulu-3-8B-SFT} & 65.4 & 53.9 & 59.9 & 69.1 & 62.1 \\
    & \texttt{Llama-3.1-Tulu-3-8B-DPO} & 76.5 & 41.9 & 58.5 & 57.5 & 58.6 \\
    & \texttt{Llama-3.1-Tulu-3-8B} & 78.5 & 38.6 & 58.2 & 59.7 & 58.8 \\
    \midrule

    \multirow{4}{*}{TII}
    & \texttt{Falcon3-7B-Instruct} & 50.6 & 57.0 & 50.5 & 74.2 & 58.1 \\
    & \texttt{Falcon3-3B-Instruct} & 70.4 & 52.4 & 57.2 & 55.3 & 58.8 \\
    & \texttt{Falcon3-1B-Instruct} & 65.4 & 44.3 & 50.4 & 59.3 & 54.8 \\
    & \texttt{Falcon3-10B-Instruct} & 53.1 & 51.5 & 57.4 & 68.8 & 57.7 \\
    \bottomrule
    \end{tabular}
  }
  \caption{\textbf{Bradley-Terry Performance.}}
  \label{tab:bt_perf}
\end{table*}

\begin{figure}[t]
  \includegraphics[width=\columnwidth]{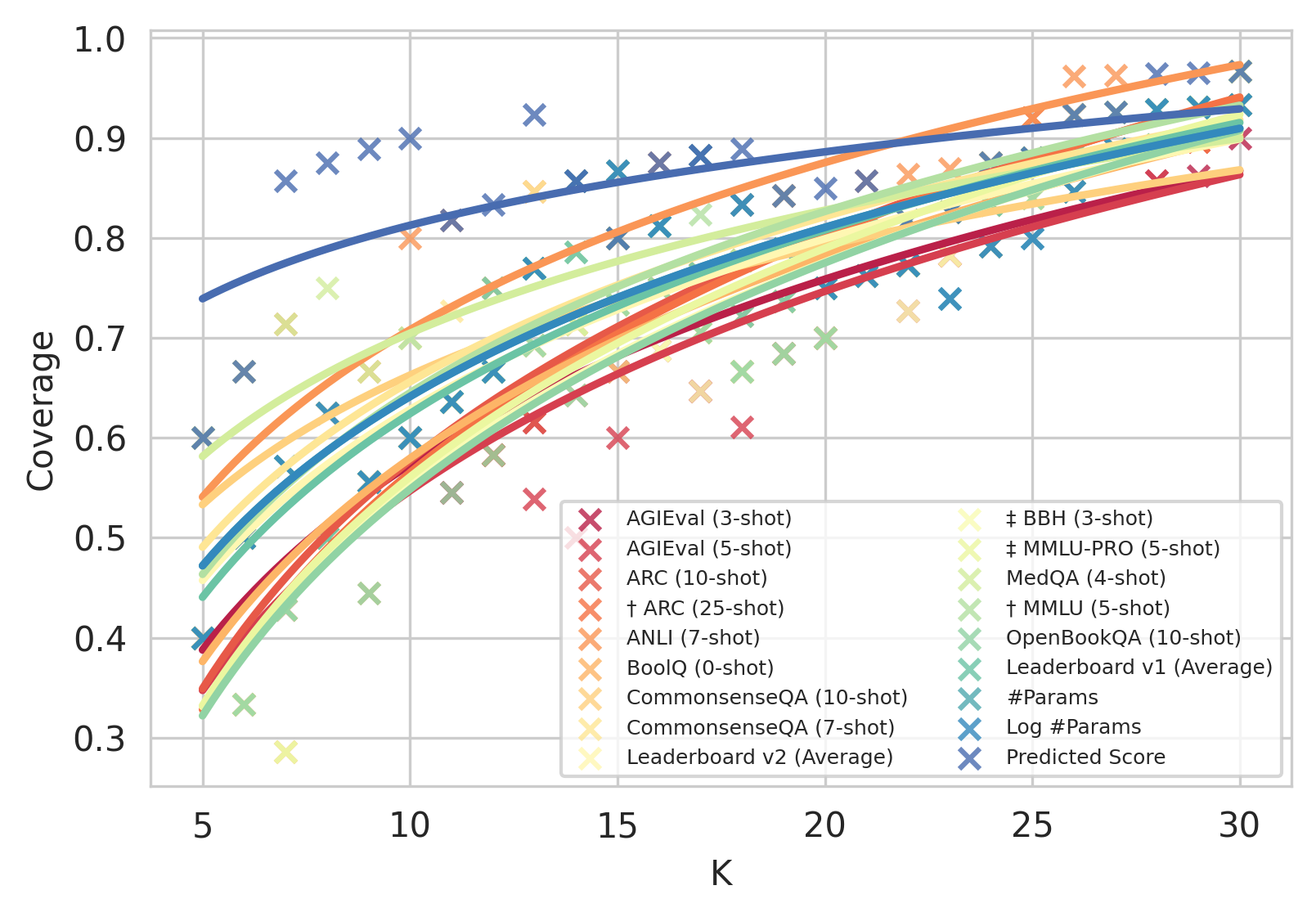}
  \caption{\textbf{Benchmarks vs. Predicted Score Coverage.} We only retain benchmarks with at least 0.4 and 0.6 coverage at $k=5$ and $k=10$, respectively.}
  \label{fig:more_cov_pred}
\end{figure}

\section{Bradley-Terry vs. Regression}

\begin{figure}[t]
  \includegraphics[width=\columnwidth]{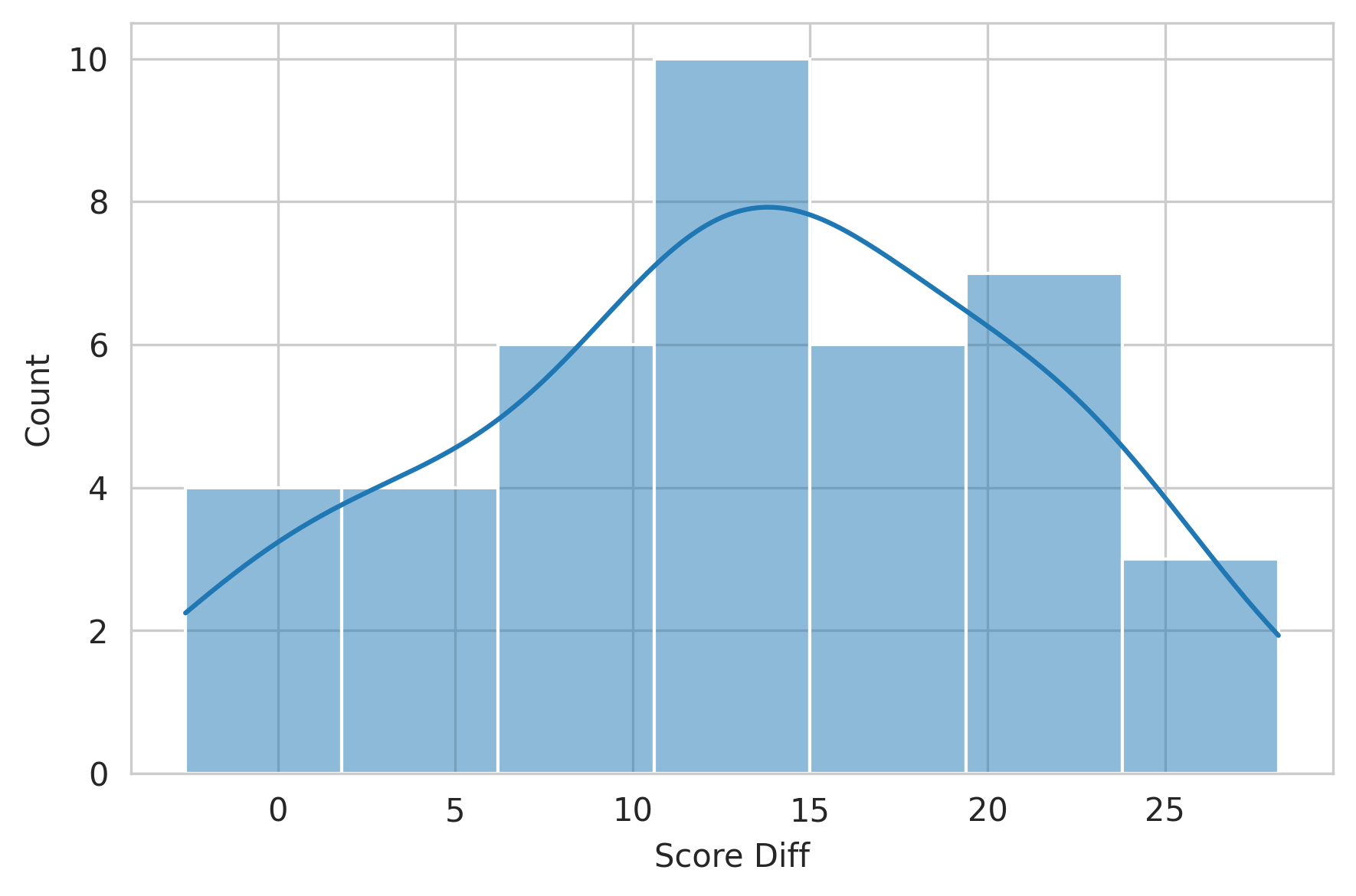}
  \caption{\textbf{Bradley-Terry vs. Regression Performance Difference.} A positive value indicates a better performance on the Regression method.}
  \label{fig:btvr}
\end{figure}

\paragraph{Setup.}
The training method is one of the early design choices for reward modeling, significantly influencing the costly data curation process, as the data format is often not easily transferable.
While previous works have briefly compared Bradley-Terry vs. Regression training~\cite{wang2024helpsteer2-p}, finding their similar performances on $\sim$70B models, our understanding of their differences is somewhat limited.
In our experiments, we use the HelpSteer2 and HelpSteer2-Preference datasets, which have the same underlying samples with different annotation styles\footnote{HelpSteer2-Preference excludes indistinguishable responses (denoted by human annotators), which Bradley-Terry w/ Binary Preferences can not model.}.
This setup presents an opportunity to compare these two approaches fairly.

\begin{figure*}[t]
  \includegraphics[width=\textwidth]{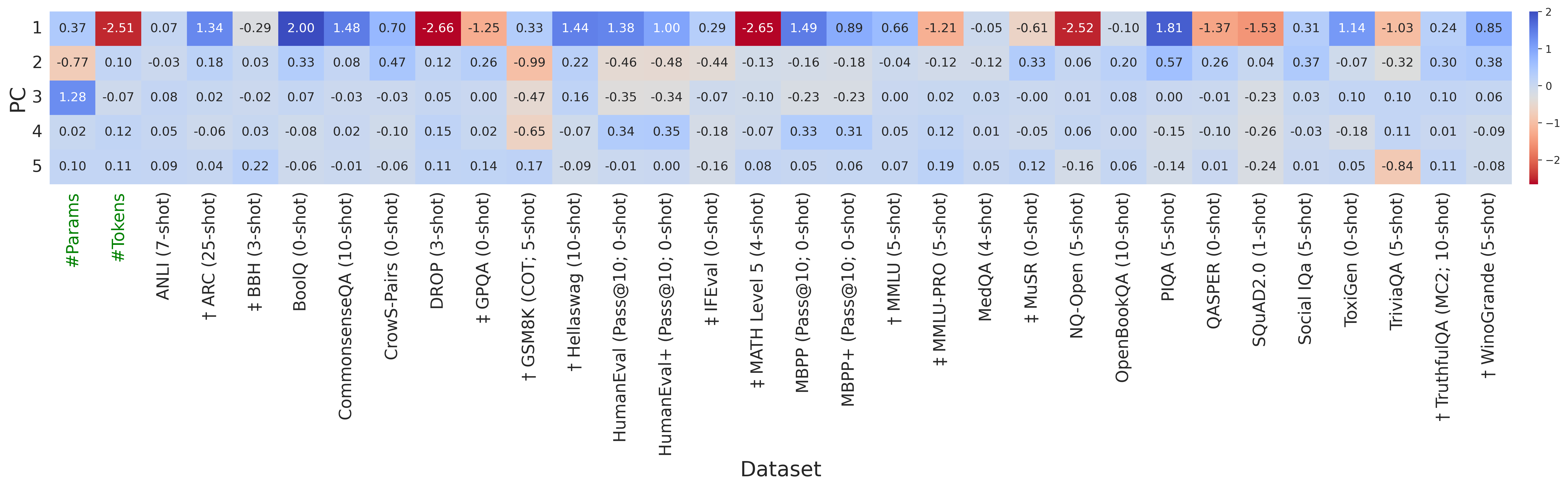}
  \caption{\textbf{Principal Component's Weights.}}
  \label{fig:pc_coeff}
\end{figure*}

\begin{figure}[t]
  \includegraphics[width=\columnwidth]{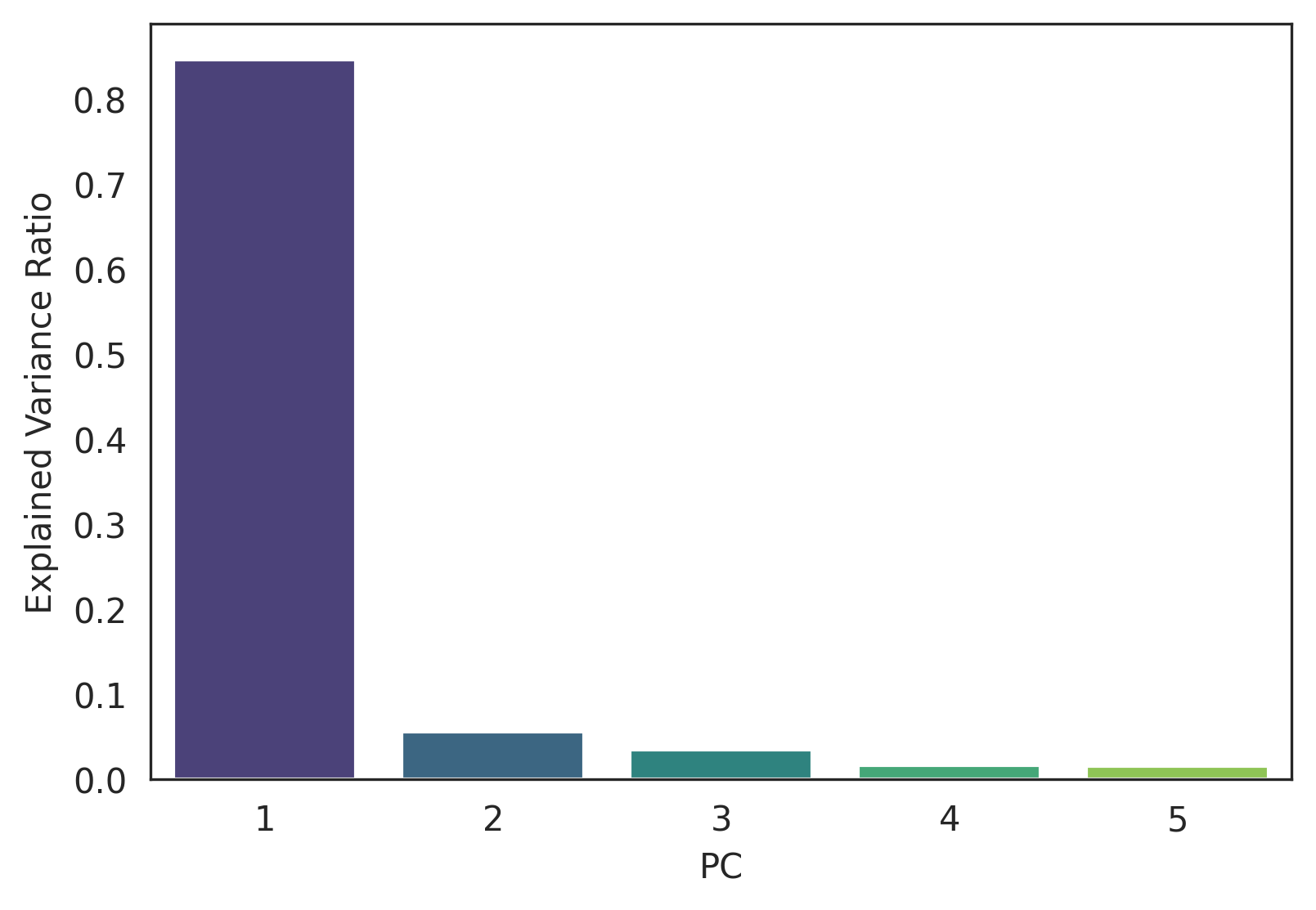}
  \caption{\textbf{PCA Explained Variance.} We find that the top 5 PCs explain $\sim$96.8\% of the variance; hence, the benchmark-model matrix is low-dimensional.}
  \label{fig:pc_var}
\end{figure}

\paragraph{Results.}
\autoref{fig:btvr} illustrates the performance difference between Bradley-Terry and Regression methods across our model pool.
As evident, the Regression models outperform their Bradley-Terry counterparts by a large margin.
We also observe that the gap is much less with stronger models (\textit{e.g.,} \texttt{Qwen2.5-7B-Instruct} and \texttt{gemma-2-9b-it}), which could lead to a performance match on 70B scale models, consistent with previous findings (see \autoref{app:full} for more details).
This observation suggests that the Regression method is less reliant on the quality of the base model, making it a better overall choice when possible.
Moreover, we note much more overfitting and instability when training with the Bradley-Terry method, making obtaining high-quality RMs more challenging.

\section{Low-dimensional Capabilities}

\paragraph{Setup.}
Prior works~\cite{ruan2024observational,polo2024sloth} have found the LLMs' capabilities to be low-dimensional, meaning that most of the variance over the standard benchmarks can be explained by a few principal components (PCs).
Since our experiments use an expanded set of benchmarks (5 vs. 32), we replicate their analysis at a larger scale.
Moreover, \citet{ruan2024observational} find that the PCs are explainable, meaning specific topics, such as reasoning or coding, can explain each of them.

\paragraph{Results.}
\autoref{fig:pc_var} illustrates the explained variance by the first five PCs ($\sim$97\%), which verifies that the benchmark-model matrix is low-dimensional.
Moreover, \autoref{fig:pc_coeff} replicates their analysis over the expanded set of benchmarks.
While some PCs showcase a strong connection to specific topics (\textit{e.g.,} PC4 $\approx$ Math + Coding), we can not assign clear-cut topics to them, in contrast to prior findings.

\section{Implementation Details}
All our experiments are carried out on a server with 8 $\times$ RTX A6000 GPUs with 48GB VRAM, 500GB RAM, and 64 CPU cores.
Moreover, we implemented our code using Hugging Face Transformers~\citep{wolf-etal-2020-transformers} and PyTorch~\citep{paszke2019pytorch} libraries.

\end{document}